\documentclass[final]{cvpr}

\usepackage{times}
\usepackage{epsfig}
\usepackage{graphicx}
\usepackage{color}
\usepackage[dvipsnames]{xcolor}
\usepackage[font=normal]{subfig}
\usepackage{tabularx}
\usepackage{booktabs}
\usepackage{amsmath}
\usepackage{amssymb}
\usepackage{multirow}
\usepackage[hang,flushmargin]{footmisc}

\usepackage[pagebackref=true,breaklinks=true,colorlinks,bookmarks=false,citecolor=blue]{hyperref}

\def\cvprPaperID{52} 
\def\confYear{CVPR 2021}

\newcommand{\tabcspace}{\vspace{-2.5mm}}
\newcommand{\tabspace}{\vspace{-4.5mm}}
\newcommand{\figcspace}{\vspace{-2.5mm}}
\newcommand{\figspace}{\vspace{-3.5mm}}

\begin{document}

\title{NTIRE 2021 Challenge on Image Deblurring}

\author{
    Seungjun Nah$^\dagger$ \and Sanghyun Son$^\dagger$ \and Suyoung Lee$^\dagger$ \and Radu Timofte$^\dagger$ \and Kyoung Mu Lee$^\dagger$    
    
    \and Liangyu Chen \and Jie Zhang \and Xin Lu \and Xiaojie Chu \and Chengpeng Chen       
    \and Zhiwei Xiong \and Ruikang Xu \and Zeyu Xiao \and Jie Huang \and Yueyi Zhang        
    \and Si Xi \and Jia Wei                                                                 
    
    \and Haoran Bai \and Songsheng Cheng \and Hao Wei \and Long Sun 
    \and Jinhui Tang \and Jinshan Pan                                                       
    \and Donghyeon Lee \and Chulhee Lee \and Taesung Kim                                    
    \and Xiaobing Wang \and Dafeng Zhang                                                    
    
    \and Zhihong Pan \and Tianwei Lin \and Wenhao Wu \and Dongliang He \and Baopu Li \and Boyun Li 
    \and Teng Xi \and Gang Zhang \and Jingtuo Liu \and Junyu Han \and Errui Ding            
    
    \and Guangpin Tao \and Wenqing Chu \and Yun Cao \and Donghao Luo \and Ying Tai \and Tong Lu 
    \and Chengjie Wang \and Jilin Li \and Feiyue Huang                                      
    \and Hanting Chen \and Shuaijun Chen \and Tianyu Guo \and Yunhe Wang                    
    
    \and Syed Waqas Zamir \and Aditya Arora \and Salman Khan \and Munawar Hayat 
    \and Fahad Shahbaz Khan \and Ling Shao                                                  
    \and Yushen Zuo \and Yimin Ou \and Yuanjun Chai \and Lei Shi                            
    \and Shuai Liu \and Lei Lei \and Chaoyu Feng                                            
    \and Kai Zeng \and Yuying Yao \and Xinran Liu                                           
    \and Zhizhou Zhang \and Huacheng Huang                                                  
    \and Yunchen Zhang \and Mingchao Jiang \and Wenbin Zou                                  
    \and Si Miao                                                                            
    \and Yangwoo Kim                                                                        
    \and Yuejin Sun                                                                         
    \and Senyou Deng \and Wenqi Ren \and Xiaochun Cao \and Tao Wang                         
    \and Maitreya Suin \and A. N. Rajagopalan                                               
    \and Vinh Van Duong \and Thuc Huu Nguyen \and Jonghoon Yim \and Byeungwoo Jeon          
    \and Ru Li \and Junwei Xie                                                              
    \and Jong-Wook Han \and Jun-Ho Choi \and Jun-Hyuk Kim \and Jong-Seok Lee                
    \and Jiaxin Zhang \and Fan Peng                                                         
    
    \and David Svitov \and Dmitry Pakulich                                                  
    \and Jaeyeob Kim \and Jechang Jeong                                                     
    
}

\maketitle

\begin{abstract}
    Motion blur is a common photography artifact in dynamic environments that typically comes jointly with the other types of degradation.
    This paper reviews the NTIRE 2021 Challenge on Image Deblurring.
    In this challenge report, we describe the challenge specifics and the evaluation results from the 2 competition tracks with the proposed solutions.
    While both the tracks aim to recover a high-quality clean image from a blurry image, different artifacts are jointly involved.
    In track 1, the blurry images are in a low resolution while track 2 images are compressed in JPEG format.
    In each competition, there were 338 and 238 registered participants and in the final testing phase, 18 and 17 teams competed.
    The winning methods demonstrate the state-of-the-art performance on the image deblurring task with the jointly combined artifacts.
\end{abstract}

{\let\thefootnote\relax\footnotetext{
\noindent
$^\dagger$ S. Nah (seungjun.nah@gmail.com, Seoul National University), S. Son, S. Lee, R. Timofte, K. M. Lee are the NTIRE 2021 challenge organizers, while the other authors participated in the challenge. 
\\Appendix~\ref{sec:appendix} contains the authors' teams and affiliations.
\\Website: \url{https://data.vision.ee.ethz.ch/cvl/ntire21/}}}

\section{Introduction}
\label{sec:intro}

Motion blur is a prevalent artifact in dynamic scene photography.
Hand-held cameras are prone to shake while the objects in the scene can move during the exposure.
Moreover, images are typically degraded from joint visual artifacts including motion blur, low resolution, compression artifacts, noise, etc.
Image deblurring aims to recover a clean image from such a degraded blurry image.

Most modern image restoration techniques including image deblurring adopt machine-learning approaches that derive knowledge from training data.
For deblurring problem, the pairs of blurry and sharp images could be obtained by synthesizing blur from high-speed videos~\cite{Nah_2017_CVPR,Su_2017_CVPR,noroozi2017motion,Shen_2019_ICCV_Human_aware,Nah_2019_CVPR_Workshops_REDS}.
Especially, REDS dataset~\cite{Nah_2019_CVPR_Workshops_REDS} is designed to generate high-quality images as well as realistic image degradation.
Recently, there were attempts to construct datasets with real blurry images by using a beam splitter~\cite{rim2020real,zhong2020efficient_estrnn} and 2 cameras.
For such hardware-based approaches, evenly splitting the brightness and precisely aligning the image pair remains an issue.

To develop and benchmark deblurring algorithms, image and video deblurring challenges were hosted in the NTIRE 2019 and 2020 workshops.
In the NTIRE 2019 Challenge, video deblurring~\cite{Nah_2019_CVPR_Workshops_Deblur} and super-resolution methods under low-resolution~\cite{Nah_2019_CVPR_Workshops_SR} were developed.
In the NTIRE 2020 Challenge, single image deblurring methods~\cite{Nah_2020_CVPR_Workshops_Deblur} are benchmarked.

Succeeding the prior challenges, NTIRE 2021 Challenge on Image Deblurring considers image deblurring problem under additional artifacts.
In track 1, the blurry images are in a lower resolution than the target resolution.
Thus, high-frequency information is more scarce in the input.
In track 2, the blurry image suffers from JPEG compression artifacts. 
In constrast to most deblurring methods that only consider pure motion blur, the joint image restoration tasks pose more challenging and practical scenario.

This challenge is one of the NTIRE 2021 associated challenges: nonhomogeneous dehazing~\cite{ancuti2021ntire}, defocus deblurring using dual-pixel~\cite{abuolaim2021ntire}, depth guided image relighting~\cite{elhelou2021ntire}, image deblurring, multi-modal aerial view imagery classification~\cite{liu2021ntire}, learning the super-resolution space~\cite{lugmayr2021ntire}, quality enhancement of heavily compressed videos~\cite{yang2021ntire}, video super-resolution~\cite{son2021ntire}, perceptual image quality assessment~\cite{gu2021ntire}, burst super-resolution~\cite{bhat2021ntire}, high dynamic range~\cite{perez2021ntire}.

\section{Related Works}
\label{sec:related_works}

We describe the deep learning based image deblurring methods as well as the super-resolution and image deblocking (decompression). 

\subsection{Image Deblurring}
\label{sec:related_deblurring}

Deep learning was applied to dynamic scene deblurring by constructing datasets with high-speed cameras~\cite{Nah_2017_CVPR,Su_2017_CVPR,noroozi2017motion}.
Multi-scale networks~\cite{Nah_2017_CVPR,Tao_2018_CVPR,Gao_2019_CVPR} followed the coarse-to-fine approaches in optimization based frameworks~\cite{levin2009understanding,cho2009fast,xu2010two,Kim_2013_ICCV,Kim_2014_CVPR}.
Motivated that motion blur is spatially varying, spatially non-uniform operations as well as convolution were adopted.
Spatially variant RNN was proposed as a deconvolution operator~\cite{Zhang_2018_CVPR_svrn} and deformable convolution~\cite{Zhu_2019_CVPR_deformable_V2} was used to approximate the shape of blur kernel~\cite{Yuan_2020_CVPR}.
In contrast to the single feed-forward computation, MTRNN~\cite{Park_2020_ECCV_MTRNN} proposed to remove partial blur multiple times with a small module.
In \cite{Shen_2019_ICCV_Human_aware}, more attention was paid to human bodies as they tend to be the main objects in photography.

On the other hand, there were efforts to optimize such models with focus on perceptual quality.
Adversarial loss~\cite{Nah_2017_CVPR,Kupyn_2018_CVPR,Kupyn_2018_CVPR}, perceptual loss~\cite{Kupyn_2018_CVPR,Kupyn_2019_ICCV} were used.
Also, unsupervised training with cycle-consistency~\cite{Lu_2019_CVPR} was attempted in domain-specific deblurring.

Specific to face and text images, \cite{Xu_2017_ICCV} proposed a joint deblurring and super-resolution model with deep learning.
To cope with the ill-posedness of the joint task, adversarial training framework is employed to learn a category-specific prior.
Later, dual branch architectures were proposed~\cite{zhang2018gated,zhang2018deep}.
In \cite{zhang2018gated}, the features from deblurring module and the super-resolution feature extraction module are fused by gate module to obtain high-resolution reconstruction result.
In contrast, \cite{zhang2018deep} uses a feature extraction module followed by the deblurring module and the high-resolution prediction module.
The auxiliary deblurring branch is used to aid train the feature extraction module.

On the other hand, little attempts were made to handle compression artifacts in deblurring task.
In case of video deblurring, MPEG compression was considered in NTIRE 2019 Challenge on Video Deblurring~\cite{Nah_2019_CVPR_Workshops_Deblur}.

\subsection{Image Super-Resolution}
\label{sec:related_sr}

From the the early CNNs for super-resolution~\cite{dong2014learning_srcnn,kim2016accurate_vdsr}, many model architectures were proposed.
Faster models were developed using sub-pixel convolutions~\cite{dong2016accelerating,Shi_2016_CVPR_espcn}.
Later, residual networks~\cite{He_2016_ECCV} were widely adopted in the later methods~\cite{Ledig_2017_CVPR,Lim_2017_CVPR_Workshops_edsr,He_2019_CVPR_oisr} as well as dense connections~\cite{Tong_2017_ICCV_srdensenet,Zhang_2018_CVPR_rdn}.
Multi-scale models were also proposed to handle information in different frequency bands~\cite{Lai_2017_CVPR,Cai_2019_ICCV_toward_real}
Back projection networks were developed to provide iterative feedback mechanism~\cite{Haris_2018_CVPR,Li_2019_CVPR_feedback}.
In order to focus on relatively more useful features, attention modules were applied to the channels~\cite{Zhang_2018_ECCV_rcan,Dai_2019_CVPR_second_order} and spatial location~\cite{Mei_2020_CVPR_cross_scale} on feature maps.
Also, high-level information were jointly used to aid super-resolution performance~\cite{Wang_2020_CVPR}.

In contrast to conventional super-resolution methods considering bicubic downsampling, kernel-based methods tried to handle general downsampling methods~\cite{Gu_2019_CVPR_ikc,Zhang_2019_CVPR_deep_plug,Zhou_2019_ICCV_kmsr}.
To make deployed super-resolution model adapt to the test image, meta-learning was applied~\cite{Hu_2019_CVPR_metasr}.











\subsection{Image Deblocking}
\label{sec:related_deblock}

Early JPEG artifacts reduction mainly relied on image filtering~\cite{list2003adaptive,yoo2014post}, transformed domain~\cite{liew2004blocking} or via optimization~\cite{yang1995projection,li2014contrast}.
Sparsity was exploited for regularization~\cite{chang2013reducing,liu2015data,liu2016random}.
More recent deep learning methods learn to suppress the artifacts by minimizing reconstruction error on training set~\cite{dong2015compression,svoboda2016compression,Yu_2018_CVPR_rl_restore,Fu_2019_ICCV}.
To reflect the compression model of JPEG compression, the loss function were calculated on the frequency domain~\cite{guo2016building,guo2017one,Yoo_2018_CVPR}.
The traditional sparse coding schele was reflected in neural networks~\cite{Fu_2019_ICCV}.

\section{NTIRE 2021 Challenge}
\label{sec:challenge}
%

We hosted the NTIRE 2021 Challenge on Image Deblurring in order to encourage the community to develop the state-of-the-art algorithms for dynamic image deblurring in the wild condition.
The main objective of the challenge is to handle motion blur under additional joint degradation artifacts.
Following the NTIRE 2019 and 2020 challenges~\cite{Nah_2019_CVPR_Workshops_Deblur,Nah_2020_CVPR_Workshops_Deblur}, we use the REDS dataset~\cite{Nah_2019_CVPR_Workshops_REDS} to measure the performance of the results.


\begin{table*}[th]
    \centering
    \subfloat[Track 1. Low Resolution]{
        \begin{tabularx}{0.495\linewidth}{l >{\centering\arraybackslash}X >{\centering\arraybackslash}X >{\centering\arraybackslash}X >{\centering\arraybackslash}r}
            \toprule
            Team & PSNR$^\uparrow$ & SSIM$^\uparrow$ & LPIPS$_\downarrow$ & Runtime \\
            \midrule
            \textbf{VIDAR} & \textbf{29.04} & \textbf{0.8416} & \textbf{0.2397} & 1.0 \\
            netai & 28.91 & 0.8246 & 0.2569 & 12.4 \\
            NJUST-IMAG & 28.51 & 0.8172 & 0.2547 & 6.4 \\
            SRC-B & 28.44 & 0.8158 & 0.2531 & 0.9 \\
            Baidu & 28.44 & 0.8135 & 0.2704 & 40.8 \\
            MMM & 28.42 & 0.8132 & 0.2685 & 14.3 \\
            Imagination & 28.36 & 0.8130 & 0.2666 & 7.3 \\
            Noah\_CVlab & 28.33 & 0.8132 & 0.2606 & 24.5 \\
            TeamInception & 28.28 & 0.8110 & 0.2651 & 0.9 \\
            ZOCS\_Team & 28.25 & 0.8108 & 0.2636 & 2.2 \\
            Mier & 28.21 & 0.8109 & 0.2646 & 17.3 \\
            INFINITY & 28.11 & 0.8064 & 0.2734 & 2.7 \\
            DMLAB & 27.87 & 0.8009 & 0.2830 & 2.3 \\
            RTQSA-Lab & 27.78 & 0.7960 & 0.2830 & 6.5 \\
            Yonsei-MCML & 27.64 & 0.7956 & 0.2730 & 1.6 \\
            SCUT-ZS & 27.61 & 0.7936 & 0.2885 & 0.3 \\
            \textit{withdrawn team} & 27.55 & 0.7935 & 0.2785 & 0.3 \\   
            Expasoft team & 27.44 & 0.7902 & 0.2850 & 1.0 \\
            \midrule
            bicubic upsampling & 24.06 & 0.6817 & 0.5120 & - \\
            \bottomrule
        \end{tabularx}
    }
    \subfloat[Track 2. JPEG artifacts]{
        \begin{tabularx}{0.495\linewidth}{l >{\centering\arraybackslash}X >{\centering\arraybackslash}X >{\centering\arraybackslash}X >{\centering\arraybackslash}r}
            \toprule
            Team & PSNR$^\uparrow$ & SSIM$^\uparrow$ & LPIPS$_\downarrow$ & Runtime \\
            \midrule
            \textbf{The Fat, The Thin} & \multirow{2}{*}{\textbf{29.70}} & \multirow{2}{*}{0.8403} & \multirow{2}{*}{0.2319} & \multirow{2}{*}{464.8} \\
            \hfill \textbf{and The Strong} & & & & \\
            Noah\_CVlab & 29.62 & 0.8397 & 0.2304 & 76.1 \\
            CAPP\_OB & 29.60 & 0.8398 & 0.2302 & 12.7 \\
            Baidu & 29.59 & 0.8381 & 0.2340 & 71.0 \\
            SRC-B & 29.56 & 0.8385 & 0.2322 & 0.8 \\
            Mier & 29.34 & 0.8355 & 0.2546 & 17.3 \\
            VIDAR & 29.33 & \textbf{0.8565} & \textbf{0.2222} & 5.3 \\
            DuLang$^{*}$ & 29.17 & 0.8325 & 0.2411 & -\\
            TeamInception & 29.11 & 0.8292 & 0.2449 & 10.1 \\
            GiantPandaCV & 29.07 & 0.8286 & 0.2499 & 2.4 \\
            Maradona & 28.96 & 0.8264 & 0.2506 & 21.4 \\
            LAB FUD$^{*}$ & 28.92 & 0.8259 & 0.2424 & -\\
            SYJ & 28.81 & 0.8222 & 0.2546 & 1.4 \\
            Dseny & 28.26 & 0.8081 & 0.2603 & 0.6 \\
            IPCV IITM & 27.91 & 0.8028 & 0.2947 & 6.4 \\
            DMLAB & 27.84 & 0.8013 & 0.2934 & 33.2 \\
            Blur Attack & 27.41 & 0.7887 & 0.3124 & 1.7 \\
            \midrule
            no processing & 24.94 & 0.7199 & 0.3265 & -\\
            \bottomrule
        \end{tabularx}
    }
    \\
    \tabcspace
    \caption{
        \textbf{NTIRE 2021 Image Deblurring Challenge results measured on the REDS~\cite{Nah_2019_CVPR_Workshops_REDS} test dataset.}
        Teams are ordered by ranks in terms of PSNR(dB).
        The running time is the average test time~(sec) taken to generate a single output image in reproduction process using 1 Quadro RTX 8000 GPU with 48GB VRAM.
        We note that the reported timing includes I/O and initialization overhead due to the difficulty in measuring pure model inference time by modifying each implementation.
    }
    \label{tab:results}
    \tabspace
\end{table*}

{\let\thefootnote\relax\footnotetext{
\noindent
$^*$ Solutions from DuLang and LAB FUD teams were not reproducible from the submitted code.
}}

\subsection{Tracks and Competitions}
\label{sec:tracks}

In this challenge, we considered commonly witnessed visual artifacts, low-resolution and the JPEG compression as well as the motion blur.
Both the degradations make the removal of motion blur to be more difficult.
The competition consists of 2 tracks: (1) Low Resolution (2) JPEG Artifacts.

\noindent \textbf{Image Deblurring Track 1. Low Resolution} aims to develop single-image deblurring methods under $\times 4$ low resolution image than the target resolution.
A joint deblurring and super-resolution task is posed.

\noindent \textbf{Image Deblurring Track 2. JPEG Artifacts} provides the blurry images under JPEG compression.
The images are compressed by $\times 4$ ratio to keep a similar degree of information loss as Track 1.

\noindent \textbf{Competitions} Both the tracks are hosted on the CodaLab competition platform.
Each participant is required to register to the CodaLab challenge tracks to access the data and submit their deblurred results.
During the development phase, the participants use their training set to develop solutions.
The online feedback on part (every 10th) of the validation data was available.
Due to the large size of the validation set, the participants were provided with the validation data ground truth for local evaluation.
At the testing phase, each team were required to submit part of the testing set results to the CodaLab server.
Parallel with the online submission, all the deblurred images and the inference code was submitted via email.

\noindent \textbf{Evaluation} The primary evaluation metric in this challenge is PSNR.
To supplement and provide additional information, SSIM~\cite{wang2004image_ssim} and LPIPS~\cite{Zhang_2018_CVPR_unreasonable} is also measured.
The running time was measured by the organizers with the code provided by the participants, checking the reproducibility of each solution.


\section{Challenge Results}
\label{sec:results}

\begin{figure*}[t!]
    \renewcommand{\wp}{0.13}
    \centering
    \subfloat{\includegraphics[width=\wp\linewidth]{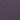}}
    \hfill
    \subfloat{\includegraphics[width=\wp\linewidth]{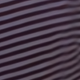}}
    \hfill
    \subfloat{\includegraphics[width=\wp\linewidth]{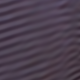}}
    \hfill
    \subfloat{\includegraphics[width=\wp\linewidth]{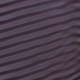}}
    \hfill
    \subfloat{\includegraphics[width=\wp\linewidth]{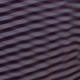}}
    \hfill
    \subfloat{\includegraphics[width=\wp\linewidth]{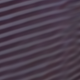}}
    \hfill
    \subfloat{\includegraphics[width=\wp\linewidth]{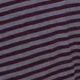}}
    \\
    \addtocounter{subfigure}{-7}
    \subfloat[Input]{\includegraphics[width=\wp\linewidth]{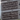}}
    \hfill
    \subfloat[]{\includegraphics[width=\wp\linewidth]{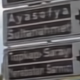}}
    \hfill
    \subfloat[]{\includegraphics[width=\wp\linewidth]{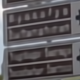}}
    \hfill
    \subfloat[]{\includegraphics[width=\wp\linewidth]{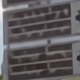}}
    \hfill
    \subfloat[]{\includegraphics[width=\wp\linewidth]{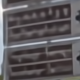}}
    \hfill
    \subfloat[]{\includegraphics[width=\wp\linewidth]{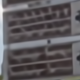}}
    \hfill
    \subfloat[GT]{\includegraphics[width=\wp\linewidth]{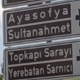}}
    \\
    \figcspace
    \caption{
        \textbf{Comparison between top-ranked results in Track 1.}
        (b) VIDAR team
        (c) netai team
        (d) NJUST-IMAG team.
        (e) SRC-B team.
        (f) Baidu team.
        Patches are cropped from REDS (test) `013/00000013' and `014/00000002', respectively.
    }
    \label{fig:comp_track1}
    \figspace
\end{figure*}
\begin{figure*}[t!]
    \renewcommand{\wp}{0.13}
    \centering
    \subfloat{\includegraphics[width=\wp\linewidth]{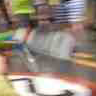}}
    \hfill
    \subfloat{\includegraphics[width=\wp\linewidth]{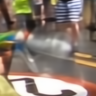}}
    \hfill
    \subfloat{\includegraphics[width=\wp\linewidth]{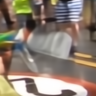}}
    \hfill
    \subfloat{\includegraphics[width=\wp\linewidth]{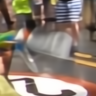}}
    \hfill
    \subfloat{\includegraphics[width=\wp\linewidth]{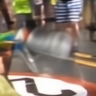}}
    \hfill
    \subfloat{\includegraphics[width=\wp\linewidth]{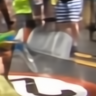}}
    \hfill
    \subfloat{\includegraphics[width=\wp\linewidth]{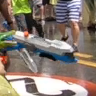}}
    \\
    \addtocounter{subfigure}{-7}
    \subfloat[Input]{\includegraphics[width=\wp\linewidth]{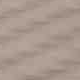}}
    \hfill
    \subfloat[]{\includegraphics[width=\wp\linewidth]{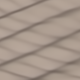}}
    \hfill
    \subfloat[]{\includegraphics[width=\wp\linewidth]{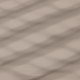}}
    \hfill
    \subfloat[]{\includegraphics[width=\wp\linewidth]{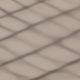}}
    \hfill
    \subfloat[]{\includegraphics[width=\wp\linewidth]{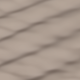}}
    \hfill
    \subfloat[]{\includegraphics[width=\wp\linewidth]{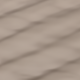}}
    \hfill
    \subfloat[GT]{\includegraphics[width=\wp\linewidth]{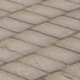}}
    \\
    \figcspace
    \caption{
        \textbf{Comparison between top-ranked results in Track 2.}
        (b) The Fat, The Thin and The Strong team.
        (c) Noah\_CVLab team.
        (d) CAPP\_OB team.
        (e) Baidu team.
        (f) SRC-B team.
        %
        %
        %
        Patches are cropped from REDS (test) `004/00000067' and `008/00000097', respectively.
    }
    \label{fig:comp_track2}
    \figspace
\end{figure*}

Each challenge track had 338 and 238 registered participants.
In each track, 18 and 17 teams submitted the results in the final testing phase.
The deblurred images were submitted along with the inference code and the trained weights for the organizers to check the reproducibility.

Table~\ref{tab:results} shows the measured performance of each team's solution as well as their inference speed.
The inference speed was measured by the organizers in a single platform.
We used Intel Xeon Gold 6248 CPU and NVIDIA Quadro RTX 8000 GPU, Samsung 860 EVO 4TB SSD.

\subsection{Architectures and Main Ideas}
There were a few novel ideas and several shared strategies between the submitted solutions.
In track 1, inspired from the video deblurring technique in EDVR~\cite{Wang_2019_CVPR_Workshops_edvr}, VIDAR and Imagination teams used pyramid deformable convolutions to align multiple features from a single image.
netai and Noah\_CVlab teams used multi-task training method to optimize features for joint deblurring and super-resolution.
Transformer architecture~\cite{vaswani2017attention} was used by Noah\_CVlab and ZOCS teams.
Non-local module~\cite{Wang_2018_CVPR_non_local} was used by NJUST-IMAG team.
In track 2, to overcome the limitation of batch normalization~\cite{ioffe2015batch}, half-instance normalization scheme was proposed by The Fat, The Thin and The Strong team.
Object edge information was exploited by Yonsei-MCML and Blur Attack teams.
Specifically to handle images with JPEG compression artifacts, CAPP\_OB team used auto-endocer loss~\cite{Kwak_2019_CVPR_Workshops}.
Dilated convolutions were adopted in many solutions to enlarge the receptive field.
Also, the attention modules were widely employed.




\subsection{Challenge Winners}
The challenge winners are determined by the PSNR scores.
In track 1, VIDAR team achieved the best restoration quality with their EDPN architecture.
They also exhibited the best SSIM and LPIPS scores in track 2.
The EDPN model is inspired from EDVR~\cite{Wang_2019_CVPR_Workshops_edvr} and exploits the similiraty information within the extracted features.
In track 2, The Fat, The Thin and The Strong team showed the best PSNR score from their proposed HINet model.
They propose Half Instance Normalization Block to design their model architecture.

\subsection{Visual Comparison}
We provide a visual comparison between the top-ranked solutions.
Figure~\ref{fig:comp_track1} shows the deblurred images from low-resolution input in Track 1.
Figure~\ref{fig:comp_track2} illustrates the images deblurred from JPEG-compressed input in Track 2.

\begin{figure*}[t]
    \centering
    \includegraphics[width=\linewidth]{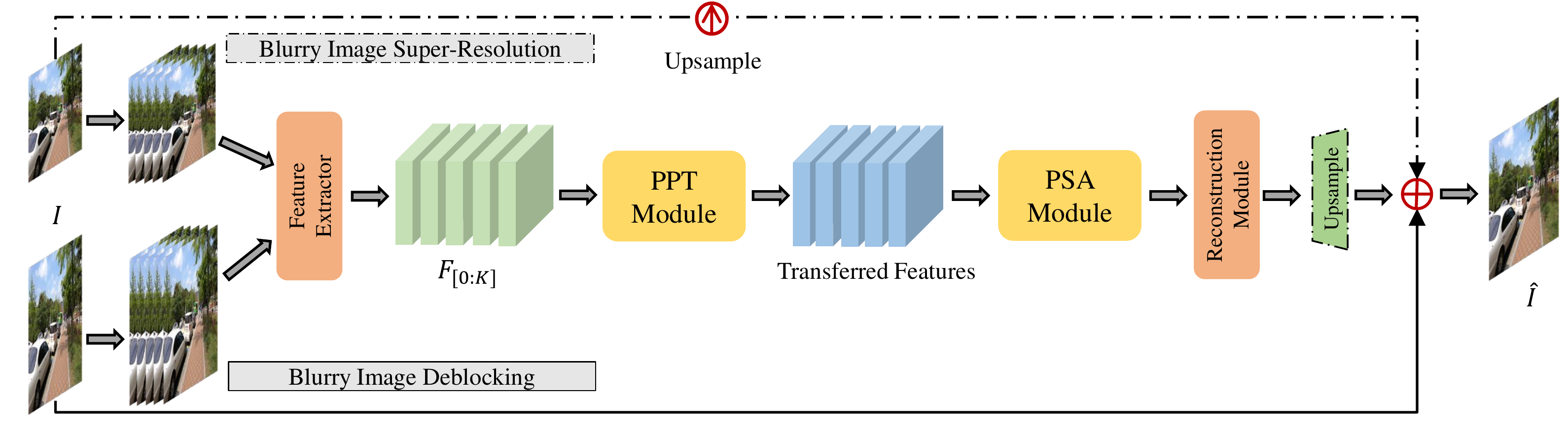}
    \\
    \figcspace
    \caption{
        \textbf{VIDAR team (Track 1 \& 2)}. Enhanced Deep Pyramid Network
    }
    \label{fig:t1_vidar}
    \figspace
\end{figure*}

\section{Challenge Methods and Teams}
\label{sec:methods}

\subsection{VIDAR}

VIDAR team proposed Enhanced Deep Pyramid Network~(EDPN)~\cite{Xu_2021_CVPR_Workshops_EDPN} for blurry image restoration from multiple degradations.
The overall structure of EDPN is shown in Figure~\ref{fig:t1_vidar}, which is inspired by EDVR~\cite{Wang_2019_CVPR_Workshops_edvr}.
Specifically, they exploit the self- and cross-scale similarities in the degraded image with two pyramid-based modules, i.e., the pyramid progressive transfer~(PPT) module and the pyramid self-attention~(PSA) module.
They first replicate the given blurry image K times (K = 4) and feed the replicated images as the input to EDPN, which aims to fully exploit the self-similarity contained in the degraded image.
The features extracted from the multiple same images by a feature extractor consisting of 18 residual blocks are fed into the PPT module.
The PPT module is designed to transfer the cross-scale similarity information from the same degraded image at the feature level with a pyramid structure, which performs the deformable convolution and generates attention masks to transfer the self-similarity information in a progressive manner.
The following PSA module is designed to aggregate information across the above transferred features, which adopts the self- and spatial-attention mechanisms to fuse the multiple features. For the blurry image super-resolution task, the fused features are fed into a reconstruction module followed by an upsampling layer.
For the blurry image deblocking task, the upsampling layer will not be necessary.
The reconstruction module is composed of 120 multi-scale residual channel attention blocks ~\cite{Zhang_2018_ECCV_rcan}.
Please refer to \cite{Xu_2021_CVPR_Workshops_EDPN} for more details.




\subsection{netai}

\begin{figure}[h]
    \centering
    \includegraphics[width=\linewidth]{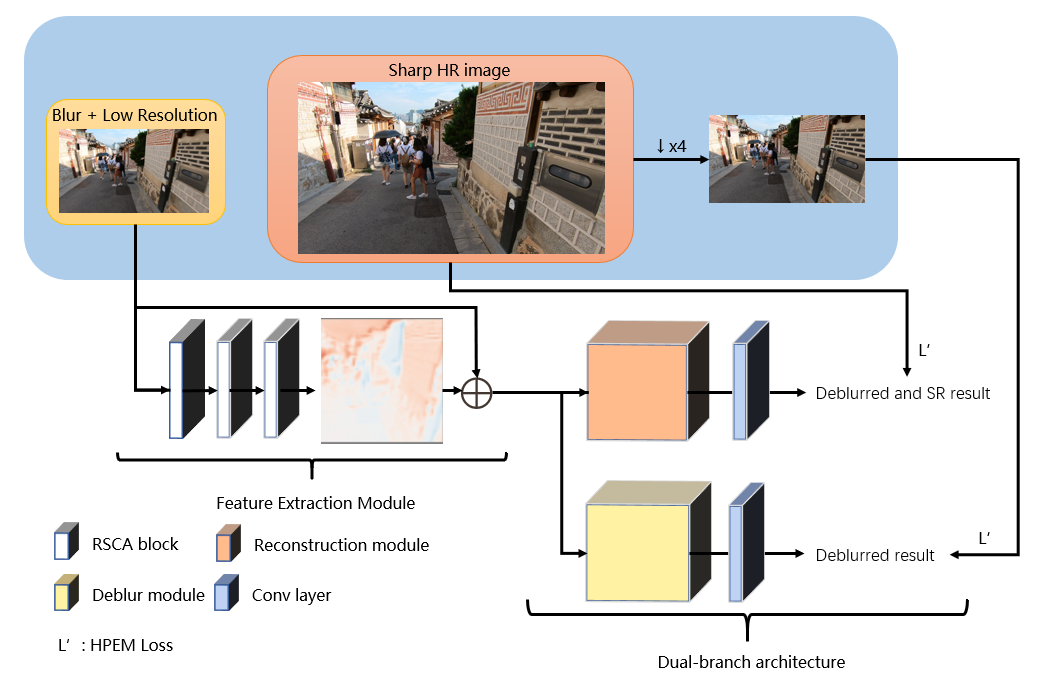}
    \\
    \figcspace
    \caption{
        \textbf{netai team (Track 1)}. Enhanced Multi-Task Network
    }
    \label{fig:t1_netai}
    \figspace
\end{figure}

netai team proposed Pixel-Guided Dual-Branch Attention Network (PDAN) for joint image deblurring and super-resolution.
The dual-branch scheme of PDAN is similar to \cite{zhang2018deep}.
In PDAN, the feature extraction module uses residual spatial and channel attention (RSCA) module, inspired by \cite{Zhang_2018_ECCV_rcan}. 
The deblurring module is a residual encoder-decoder model to enlarge the receptive field, activated by LeakyReLU layers~\cite{maas2013rectifier}. 
The shallow feature from the feature extraction module is fed into the reconstruction module to increase the spatial resolution.
The upscaling is done by scale 4 through a convolutional layer to reconstruct the HR output image. 
netai team proposed an HPEM loss function for using a hard example mining strategy to focus on the difficult areas automatically.
The whole model is jointly trained from scratch using L1 loss and then fine-tuned with the weighted sum of L1 loss and the HPEM loss.
The overall architecture is shown in Figure~\ref{fig:t1_netai}.
Please refer to \cite{xiandwei2021pdan} for more details.


\subsection{NJUST-IMAG}

\begin{figure}[h]
    \centering
    \includegraphics[width=\linewidth]{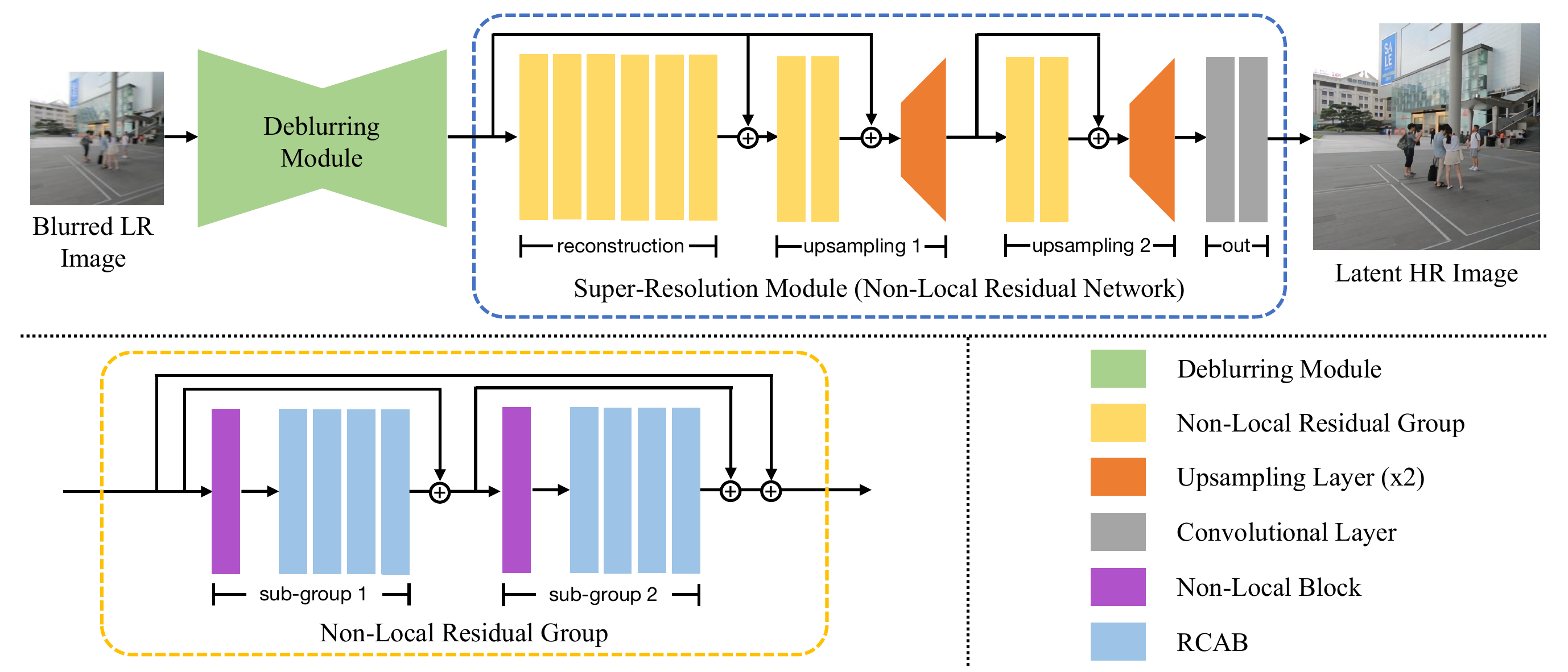}
    \\
    \figcspace
    \caption{
        \textbf{NJUST-IMAG team (Track 1)}. Learning A Cascaded Non-Local Residual Network for Super-resolving Blurry Images
    }
    \label{fig:t1_njust}
    \figspace
\end{figure}

NJUST-IMAG team developed an end-to-end network consisting of a deblurring module and a subsequent super-resolution module.
A non-local residual network~(NLRN) is proposed as the super-resolution module to better generate high-quality images.
In the NLRN, the non-local residual group is adopted as the basic unit.
The non-local residual group contains two sub-groups that each consist of a non-local block~\cite{Wang_2018_CVPR_non_local} and four RCABs~\cite{Zhang_2018_ECCV_rcan}.
The non-local architecture is effective at modeling the global information which is able to help remove the residual blur and further improve the super-resolution performance.
Self-attention is adopted to explore the relation between each image patch.
Multi-head mechanism~\cite{vaswani2017attention} is used to make the non-local block focus on more diverse global correlation.

The whole model is jointly trained starting from the pretrained deblurring module.
L1 loss and the image gradient loss are employed to train the model.
The overall architecture is shown in Figure~\ref{fig:t1_njust}.
More information can be found in \cite{bai2021cascaded}.

\subsection{SRC-B}

\begin{figure}[h]
    \centering
    \includegraphics[width=\linewidth]{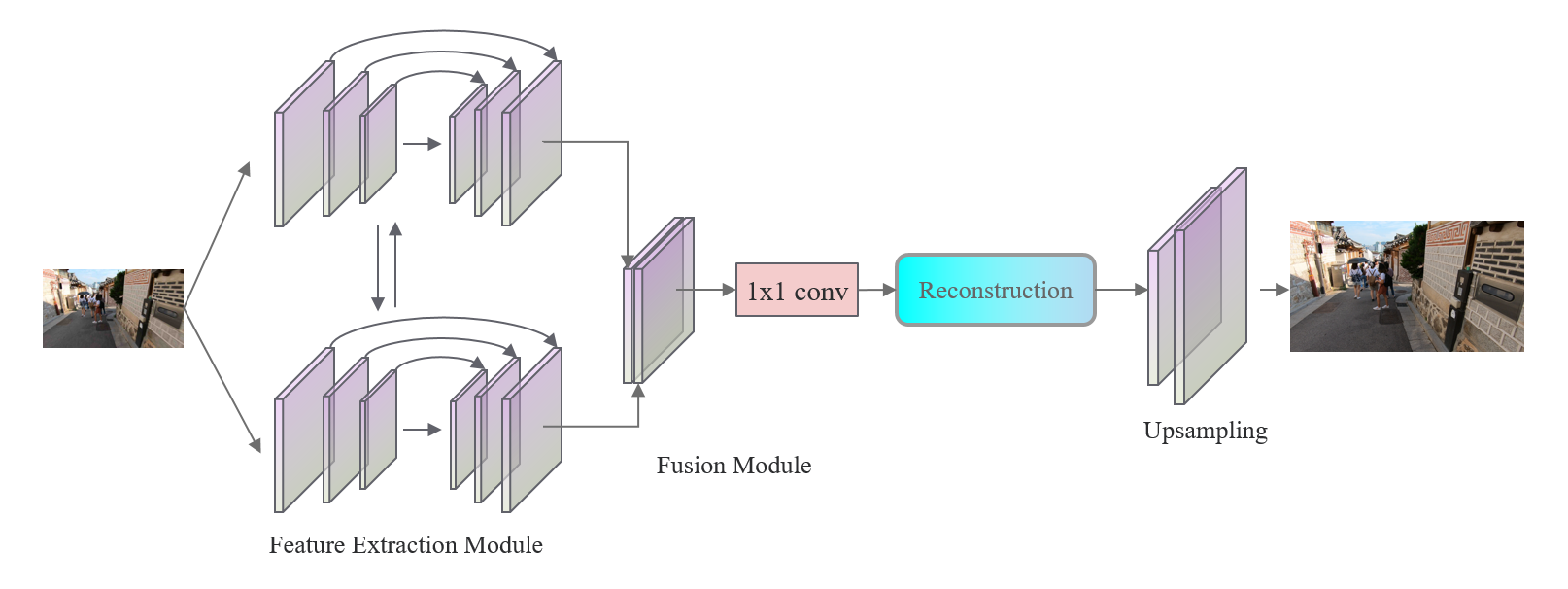}
    \\
    \figcspace
    \caption{
        \textbf{SRC-B team (Track 1 \& 2)}. MRNet: Multi-Refinement Network
    }
    \label{fig:t1_src}
    \figspace
\end{figure}

SRC-B team proposed a Multi-Refinement Network~(MRNet) for image deblurring.
MRNet was originally developed for defocus deblurring on images from dual camera and applied to single image deblurring in this competition.
MRNet is composed of 4 modules: feature extraction, fusion, reconstruction, and upsampling.
The feature extraction module computes Siamese feature from the single input image.
The features are concatenated in channel dimension and then fused by $1\times1$ convolution.
Inspired by MMDM~\cite{Liu_2020_CVPR_Workshops}, Residual Block Module~(RBM) is proposed.
RBM adopts the same configuration as MMDM, consisting of 10 residual modules and a global residual connection.
To avoid the increment in computational complexity, channel attention is not used.
Similarly to FERM in \cite{Liu_2020_CVPR_Workshops}, 5 RBM modules are used to form residual group module~(RGM).
Multi-scale RGM~(MSRGM) is constructed from the RGMs by computing parallel features with encoder-decoder structure.
Finally, the reconstruction module is composed of multiple MSRGM modules, connected in series.
On the idea that each block refines the features from the previous layer, every module has a global residual connection.

With the proposed architecture self-ensemble did not consistently increase PSNR.
Thus, multi-model ensemble strategy was used to make final results.
The overall architecture is shown in Figure~\ref{fig:t1_src}.

\subsection{Baidu}

\begin{figure}[h]
    \centering
    \includegraphics[width=\linewidth]{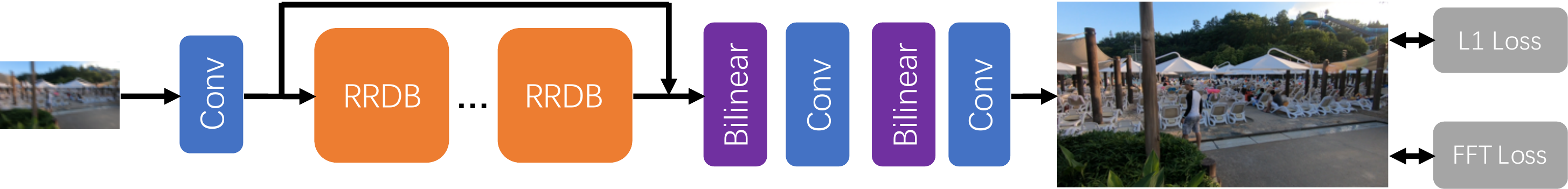}
    \\
    \figcspace
    \caption{
    \textbf{Baidu team (Track 1 \& 2)}. Joint Super-Resolution and Deblurring Using Dual Model Ensemble
    }
    \label{fig:t1_baidu}
    \figspace
\end{figure}

Baidu team proposes to improve MPRNet~\cite{zamir2021multi} and a RRDB-based model~\cite{Wang_2018_ECCV_Workshops_esrgan} and exploit the virtue of 2 models via ensemble.
The MPRNet is enhanced by adding an upsampler in the 3rd stage and by introducing an iterative process in the SAM module.
RRDB model was pretrained from DF2K dataset, combining DIV2K~\cite{Agustsson_2017_CVPR_Workshops_div2k} and Flickr2K~\cite{Timofte_2017_CVPR_Workshops_ntire2017,Lim_2017_CVPR_Workshops_edsr} datasets as \cite{Wang_2018_ECCV_Workshops_esrgan}.

Each model was trained with L1, FFT, and MS-SSIM loss with large patches of size $320\times 320$.
For the enhanced MPRNet, $640\times640$ patches were used.
The final output is generated from the ensemble of each model output that is an self-ensemble~\cite{Timofte_2016_CVPR} result from 8 geometric transforms.
The learning rate is initialized as $1\times 10^{-4}$ and halved at 20k, 30k, 35k iterations.
Adam optimizer was used.
The overall architecture is shown in Figure~\ref{fig:t1_baidu}.

\subsection{MMM}

\begin{figure}[h]
    \centering
    \includegraphics[width=\linewidth]{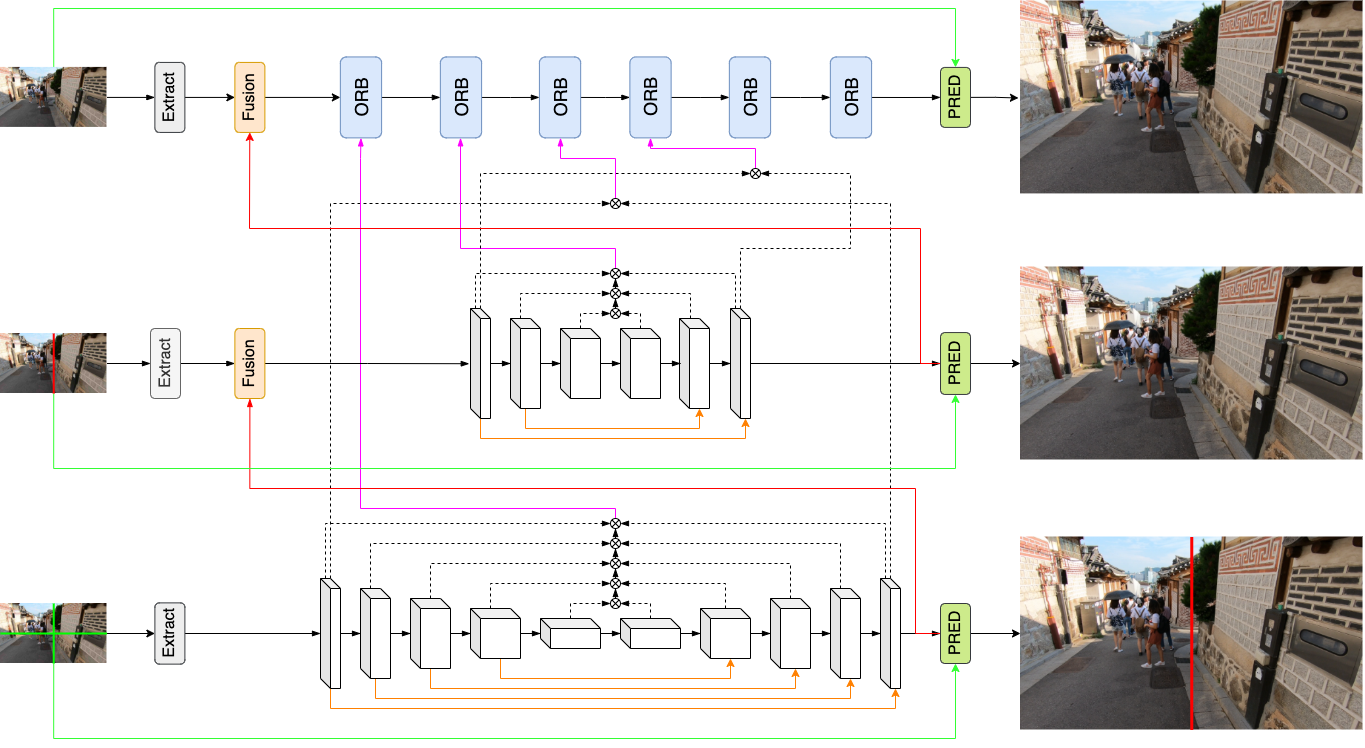}
    \\
    \figcspace
    \caption{
        \textbf{MMM team (Track 1)}. M3Net: Multi-stage, Multi-patch, and Multi-resolution
    }
    \label{fig:t1_mmm}
    \figspace
\end{figure}

MMM team proposed a M3Net model using multi-stage, multi-patch, multi-resolution strategy.
The model is divided into 3 levels and input of the each layer is a downsampled low-resolution input split in non-overlapping patches.
For the lower two stages, encoder-decoder architecture with different depth is employed to extract features of multiple scales.
The top stage does not have such encoder-decoder structure to preserve the spatial high-frequency information.
In the encoder-decoder structure, the features at the same resolution are aggregated by concatenation and convolution.
They are progressively fused with the upper stage.
CAB and ORB modules in \cite{zamir2021multi} are applied at each stage to extract features.
The skip connection between the encoders and decoders and the global skip connection with 4 times upsampling are introduced to enhance the image restoration quality.
The overall architecture is shown in Figure~\ref{fig:t1_mmm}.

\subsection{Imagination}

\begin{figure}[h]
    \centering
    \includegraphics[width=\linewidth]{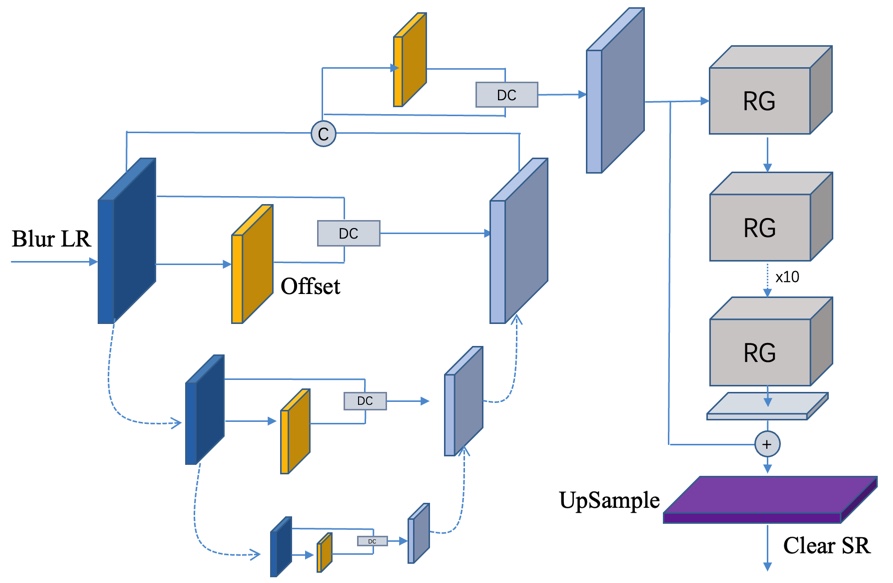}
    \\
    \figcspace
    \caption{
        \textbf{Imagination team (Track 1)}. Pyramid Deformable Convolution
    }
    \label{fig:t1_imagination}
    \figspace
\end{figure}

Imagination team proposed a pyramid deformable convolution method.
Bowworing the idea of PCD alignment module in EDVR~\cite{Wang_2019_CVPR_Workshops_edvr}, they used pyramid cascading DCN to further align individual image features.
For the refined bicubic LR aligned features, an RCAN model~\cite{Zhang_2018_ECCV_rcan} with 10 residual groups with 20 RCABs are applied.
REDS120fps dataset~\cite{Nah_2019_CVPR_Workshops_REDS} is used to synthesize additional training data.
L2 loss is used at training.
In the testing phase, 6 independent models with $\times8$ self-ensemble is used to obtain additional gains in PSNR.
The overall architecture is shown in Figure~\ref{fig:t1_imagination}.

\subsection{Noah\_CVlab}

\begin{figure}[h]
    \centering
    \includegraphics[width=\linewidth]{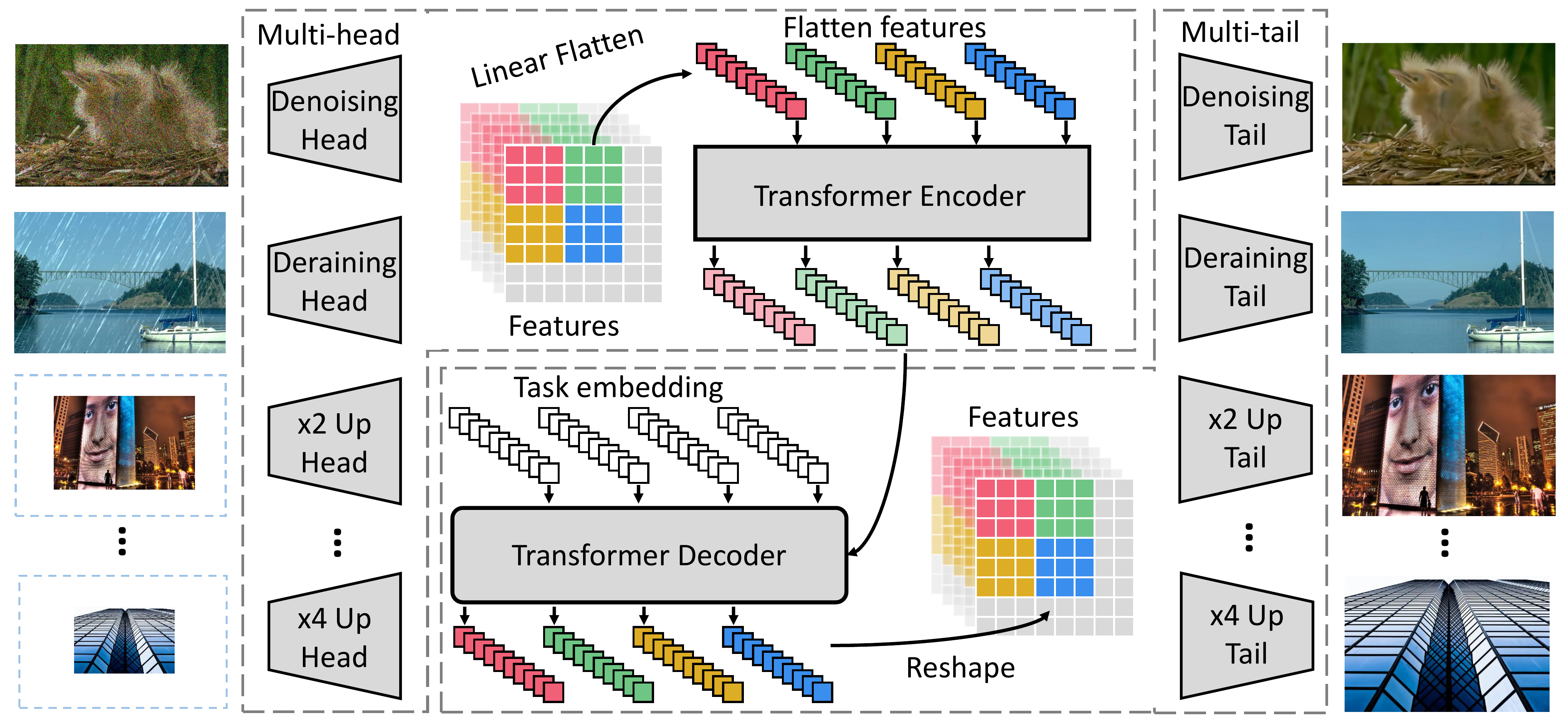}
    \\
    \figcspace
    \caption{
        \textbf{Noah\_CVlab team (Track 1 \& 2)}. Pre-trained Image Processing Transformer
    }
    \label{fig:t1_noah}
    \figspace
\end{figure}

Noah\_CVlab team adopted an Image Processing Transformer~(IPT) approach proposed in \cite{chen2020pre}.
The IPT model consists of multi-head and multi-tail for different tasks and a shared transformer body including an encoder and a decoder.
The input image is first converted to visual features and then divided into patches as visual words for subsequent processing.
The resulting image with high visual quality is reconstructed by ensembling output patches.

In the pretraining phase, there are 6 heads and tails corresponding to the six image-to-image tasks including super-resolution with scale 2, 3, 4, denoising with noise level 30 and 50, and deraining.
In the fine-tuning phase, the head and tail for $\times4$ super-resolution is chosen and the other heads and tails are dropped.
Both the heads and tails are convolutional layers.
The body consists of a 12-layer transformer encoder and a 12-layer transformer decoder.

The model is pretrained with ImageNet~\cite{deng2009imagenet} dataset and fine-tuned with GOPRO~\cite{Nah_2017_CVPR} and REDS~\cite{Nah_2019_CVPR_Workshops_REDS} datasets.
ImageNet data is utilized for generating degraded images by downsampling, adding Gaussian noise, rain streaks.
The overall architecture is shown in Figure~\ref{fig:t1_noah}.

\subsection{TeamInception}

\begin{figure}[h]
    \centering
    \includegraphics[width=\linewidth]{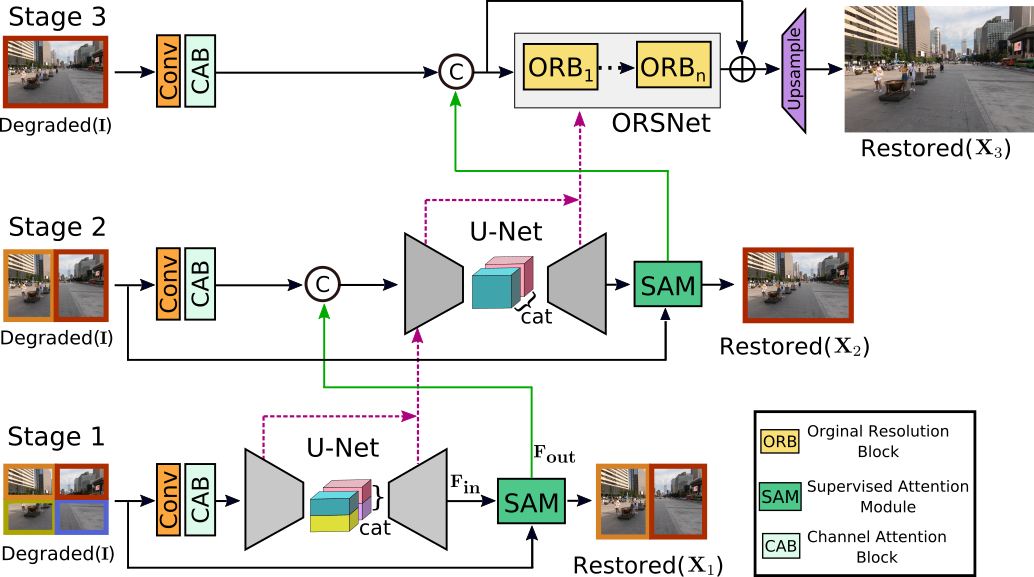}
    \\
    \caption{
        \textbf{TeamInception (Track 1 \& 2)}. Multi-Stage Progressive Image Restoration
    }
    \label{fig:t1_teaminception}
    \figspace
\end{figure}

TeamInception presented MPRNet architecture introduced in \cite{zamir2021multi}.
MPRNet consists of three stages to progressively restore images.
The first two stages are based on encoder-decoder subnetworks that learn the broad contextual information due to the large receptive field.
The last stage employs a subnetwork, ORSNet containing multiple ORB modules.
Supervised attention module~(SAM) is incorporated between the stages.
Cross-scale feature fusion mechanism is introduced where the intermediate multi-scale contextual features of the earlier subnetworks help consolidating the intermediate features of the latter subnetwork.
L1, MS-SSIM, VGG loss are used to train the model.
The overall architecture is shown in Figure~\ref{fig:t1_teaminception}.

\subsection{ZOCS\_Team}

\begin{figure}[h]
    \centering
    \includegraphics[width=\linewidth]{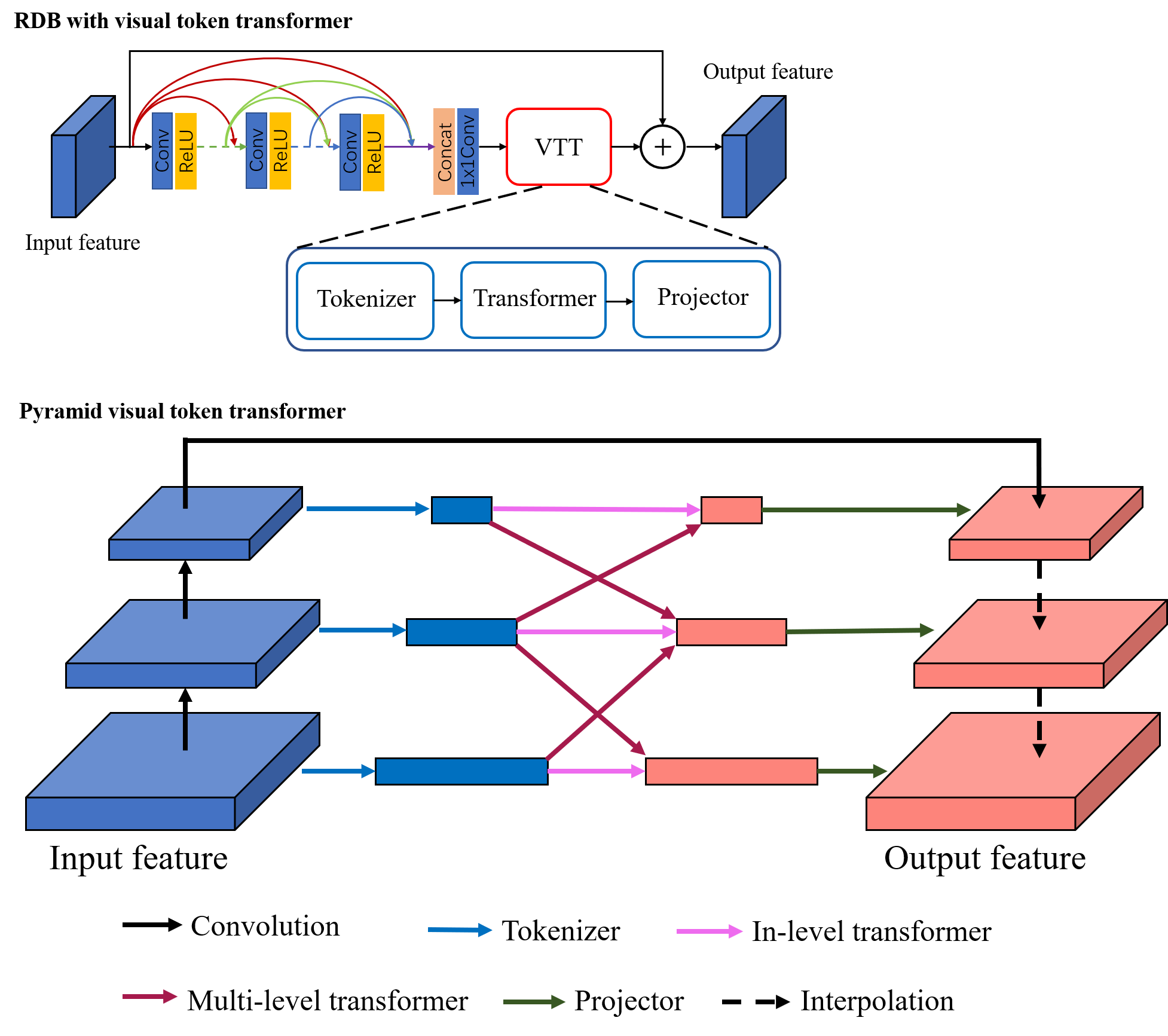}
    \\
    \figcspace
    \caption{
        \textbf{ZOCS\_Team (Track 1)}. PyVTRDN - Equip RDN with Pyramid token based transformer
    }
    \label{fig:t1_zocs}
    \figspace
\end{figure}

ZOCS\_Team used RDN~\cite{Zhang_2018_CVPR_rdn} as a baseline and added token-based transformer to the building block, RDB.
The advantages of token-based transformer can be listed as follows: 1) similar patterns in an image are grouped 2) transformers use the non-local self-similarity based on image tokens 3) less computation cost is reuiqred compared with a non-local layer.
The token-based transformer module is added after the last convolution layer of RDB.
Then pyramid token visual transformer is added after the upsampling layer of RDN.
The model is pretrained on DF2K dataset (DIV2K + Flickr2K)~\cite{Agustsson_2017_CVPR_Workshops_div2k,Lim_2017_CVPR_Workshops_edsr} is used and then fine-tuned on the REDS dataset~\cite{Nah_2019_CVPR_Workshops_REDS}.
The overall architecture is shown in Figure~\ref{fig:t1_zocs}.

\subsection{Mier}

\begin{figure}[h]
    \centering
    \includegraphics[width=\linewidth]{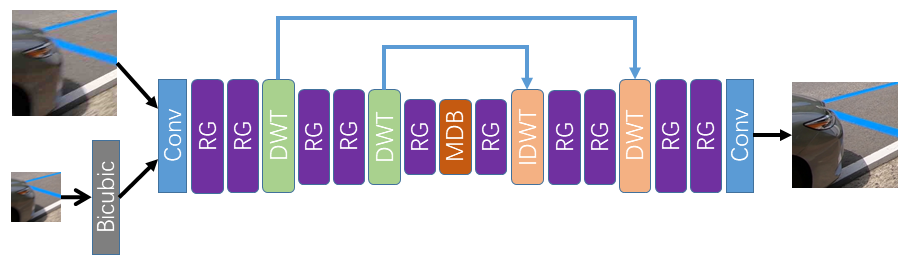}
    \\
    \figcspace
    \caption{
        \textbf{Mier team (Track 1 \& 2)}. Big UNet for Image Restoration
    }
    \label{fig:t1_mier}
    \figspace
\end{figure}

Mier team proposed a Big UNet based on MWCNN~\cite{Liu_2018_CVPR_Workshops} and RCAN~\cite{Zhang_2018_ECCV_rcan}.
They replaced the convolutional layers in MWCNN with the residual group in RCAN to enhance toe reconstruction quality.
In order to further expand the receptive field, they added a multi-scale dilated block~(MDB) from DAVANet~\cite{Zhou_2019_CVPR_davanet}.
For track 1, bicubic upsampling is applied to the input to match the image resolution.
$\times8$ self-ensemble is applied.
The overall architecture is shown in Figure~\ref{fig:t1_mier}.

\subsection{INFINITY}

INFINITY team used EDSR~\cite{Lim_2017_CVPR_Workshops_edsr} to deblur images in Track 1.
Self-ensemble was used with 8 geometric transforms.
They tested WDSR~\cite{yu2018wide}, RFDN~\cite{liu2020residual}, DRN~\cite{Guo_2020_CVPR_closed_loop}, RCAN~\cite{Zhang_2018_ECCV_rcan}, RCAN with pixel attention~\cite{zhao2020efficient} and self-calibrated convolutions~\cite{Liu_2020_CVPR_self_calibrated} and chose EDSR for better accuracy.

\subsection{DMLAB}

\begin{figure}[h]
    \centering
    \includegraphics[width=\linewidth]{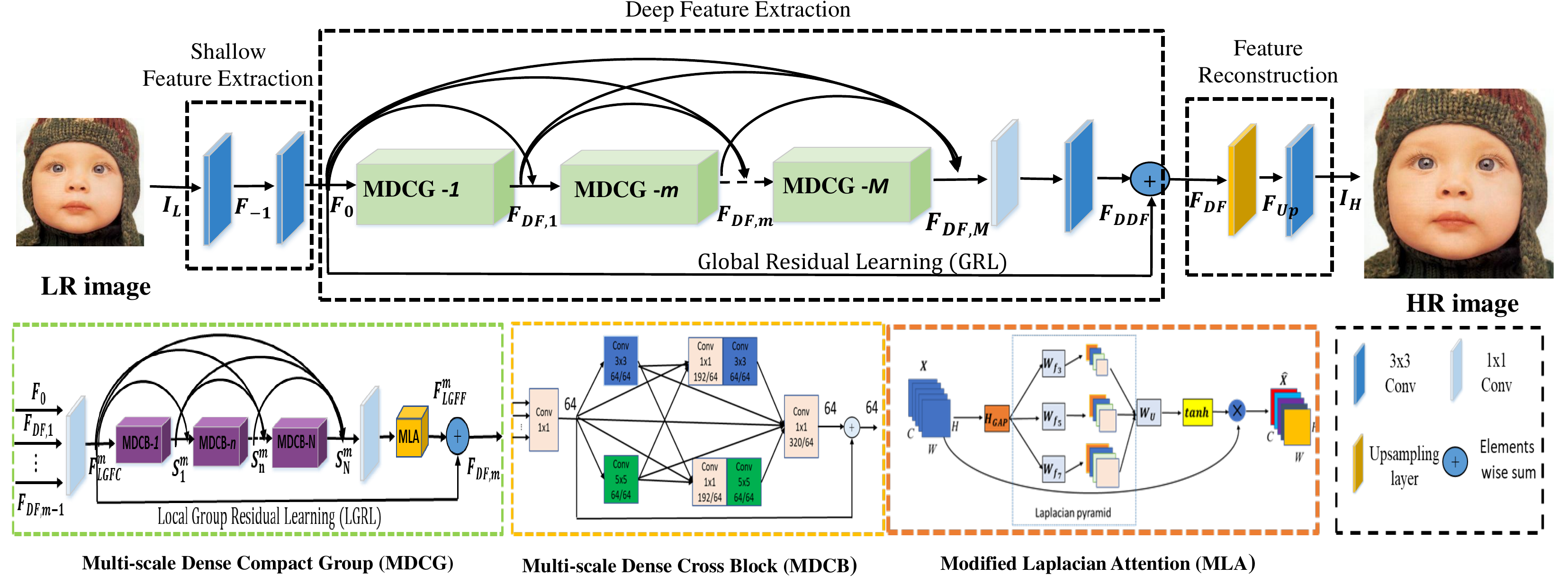}
    \\
    \figcspace
    \caption{
        \textbf{DMLAB team (Track 1 \& 2)}. Multi-scale Hierarchical Dense Residual Network
    }
    \label{fig:t1_dmlab}
    \figspace
\end{figure}

DMLAB team proposed Multi-scale Hierarchical Dense Residual Network~(MS-HDRN).
Hierarchical dense residual learning is proposed via multi-level dense connections and multi-level residual connections.
To implement multi-level dense connection, $1\times1$ convolution layers are inserted as the first and the last layers of MDCG and MDCB modules~\cite{li2020mdcn}, reducing the number of feature maps.
Inspired by \cite{Brehm_2020_CVPR_Workshops}, multi-scale feature extraction modules are used without reducing spatial resolution.
To achieve the implementation principle, MDCB~\cite{li2020mdcn} and Laplacian attention~\cite{anwar2020densely} modules are used with modifications.
The overall architecture is shown in Figure~\ref{fig:t1_dmlab}.

\subsection{RTQSA-Lab}

RTQSA-Lab team proposed Enhanced Attention Network for the competition track 1.
They presented a new attention network consisting of a Global Attention Module~(GAM) and a Local Attention Module~(LAM) to model the dependencies between layers, channels, and positions.
Specifically, the proposed GAM adaptively emphasizes hierarchical features by considering correlations among layers.
Meanwhile, LAM learns the confidence at all positions of each channel, selectively capturing more informative features.

\subsection{Yonsei-MCML}

\begin{figure}[h]
    \centering
    \includegraphics[width=\linewidth]{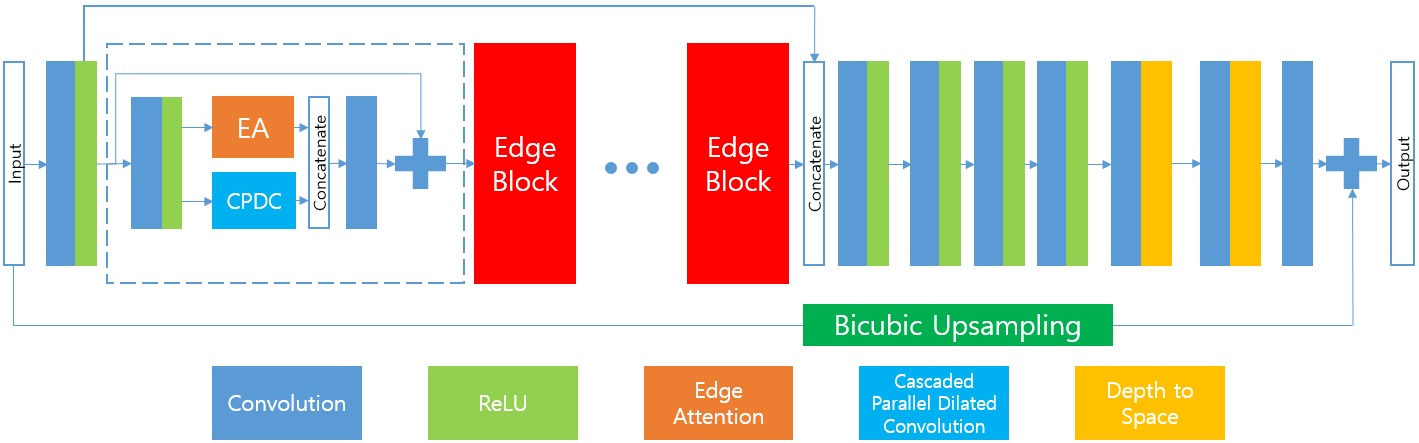}
    \\
    \figcspace
    \caption{
        \textbf{Yonsei-MCML team (Track 1)}. Edge Attention Network
    }
    \label{fig:t1_yonsei}
    \figspace
\end{figure}

Yonsei-MCML team proposed an edge detection-based attention network for image deblurring. 
On top of BA-Net~\cite{tsai2021banet} using dilated convolutions, attention module is added.
The edge information is fed into the model by Sobel filter in the horizontal and the vertical directions.
Following ESPCN~\cite{Shi_2016_CVPR_espcn}, sub-pixel convolution is employed with modification.
The overall architecture is shown in Figure~\ref{fig:t1_yonsei}.

\subsection{SCUT-ZS}

SCUT-ZS team applied EDSR~\cite{Lim_2017_CVPR_Workshops_edsr} in image deblurring task in participation to track 1.


\subsection{Expasoft team}

\begin{figure}[h]
    \centering
    \includegraphics[width=\linewidth]{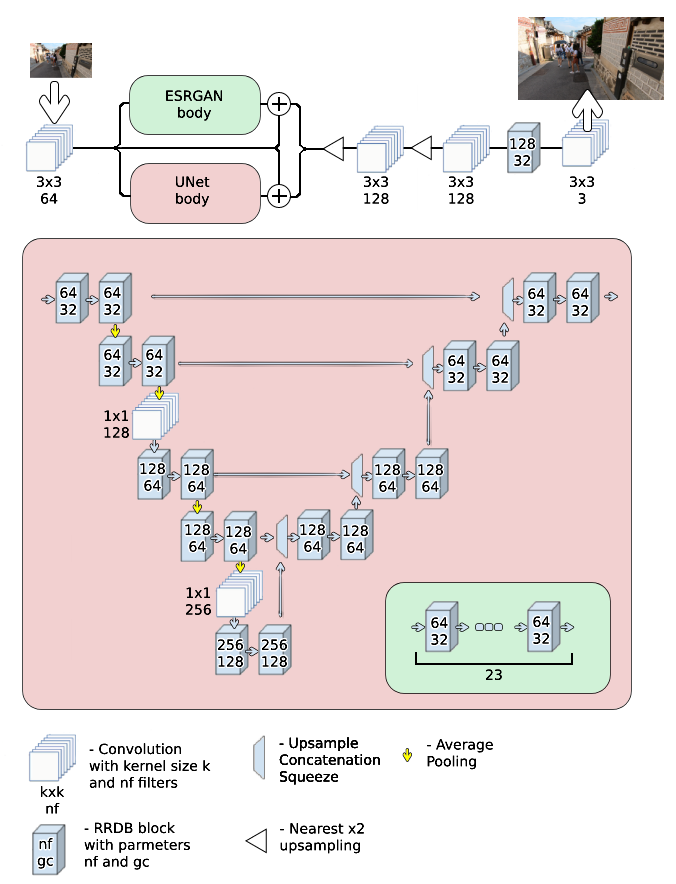}
    \\
    \figcspace
    \caption{
        \textbf{Expasoft team (Track 1)}. BowNet
    }
    \label{fig:t1_expasoft}
    \figspace
\end{figure}

Expasoft team proposed a BowNet architecture, combining ESRGAN~\cite{Wang_2018_ECCV_Workshops_esrgan} and UNet~\cite{ronneberger2015u}.
The presented UNet consists of RRDB blocks~\cite{Wang_2018_ECCV_Workshops_esrgan} where the changes in the number of channels are made by $1\times1$ convolutions.
Average pooling is used to reduce the scale of feature maps in UNet.
The UNet and the ESRGAN body are applied in parallel and the extracted features are concatenated and fused in the next layers.
The resolution is increased by a sequence of $3\times3$ convolutions and nearest-neighbor upsampling.
The overall architecture is shown in Figure~\ref{fig:t1_expasoft}.

\subsection{The Fat, The Thin and The Strong}

\begin{figure}[h]
    \centering
    \includegraphics[width=\linewidth]{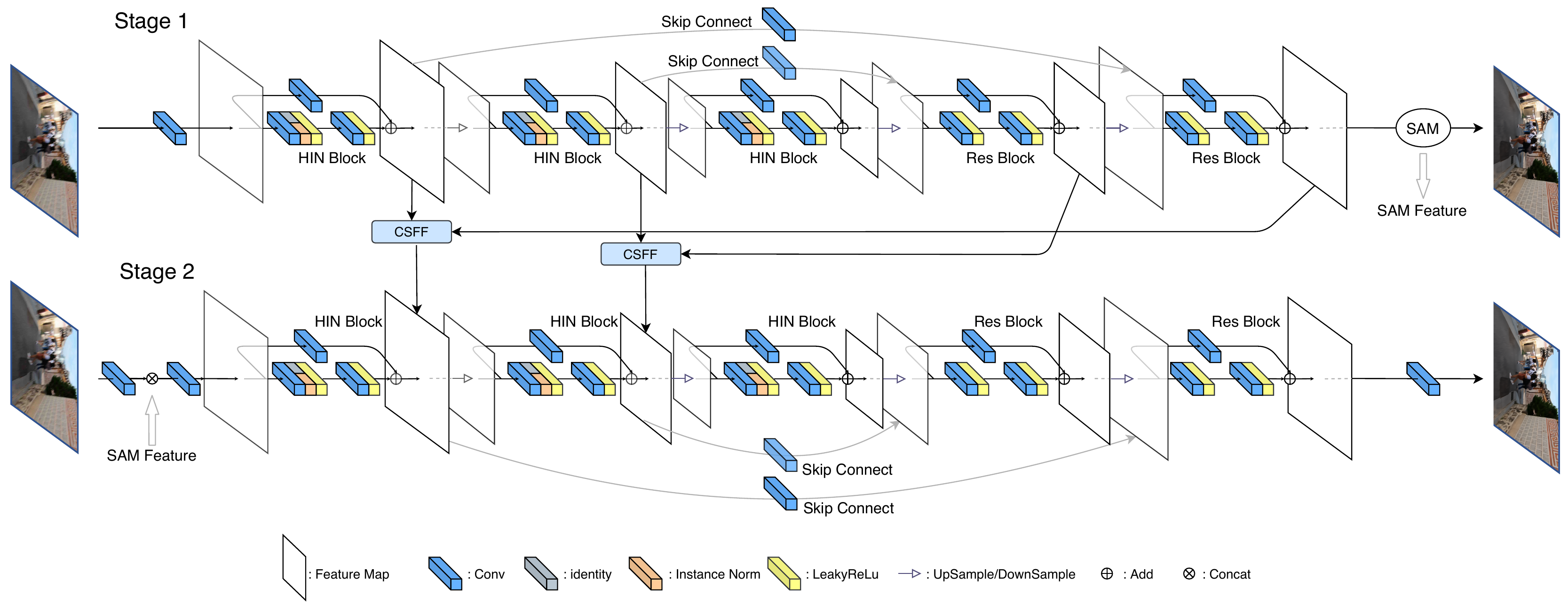}
    \\
    \figcspace
    \caption{
        \textbf{The Fat, The Thin and The Strong team (Track 2)}. HINet: Half Instance Normalization Network for Image Restoration
    }
    \label{fig:t2_fts}
    \figspace
\end{figure}

The Fat, The Thin, and The Strong team proposes a two-stage feature completion network.
Five feature representation for each stage is presented to effectively increase the receptive field.
At each stage, a convolutional feature is extracted followed by a body architecture similar to UNet~\cite{ronneberger2015u}.
To replace the batch normalization, half-IN block is designed in the encoding stage.
Half-IN block uses both the non-normalized and the normalized feature from instance normalization~\cite{ulyanov2016instance}.
SAM block from MPRNet~\cite{zamir2021multi} is adopted to refine feature and interact with the input features of the second stage.
3 models were used for ensemble but not much PSNR boost were observed in the REDS validation set.
The overall architecture is shown in Figure~\ref{fig:t2_fts}.
Please refer to \cite{chen2021hinet} for more details.

\subsection{CAPP\_OB}

\begin{figure}[h]
    \centering
    \includegraphics[width=\linewidth]{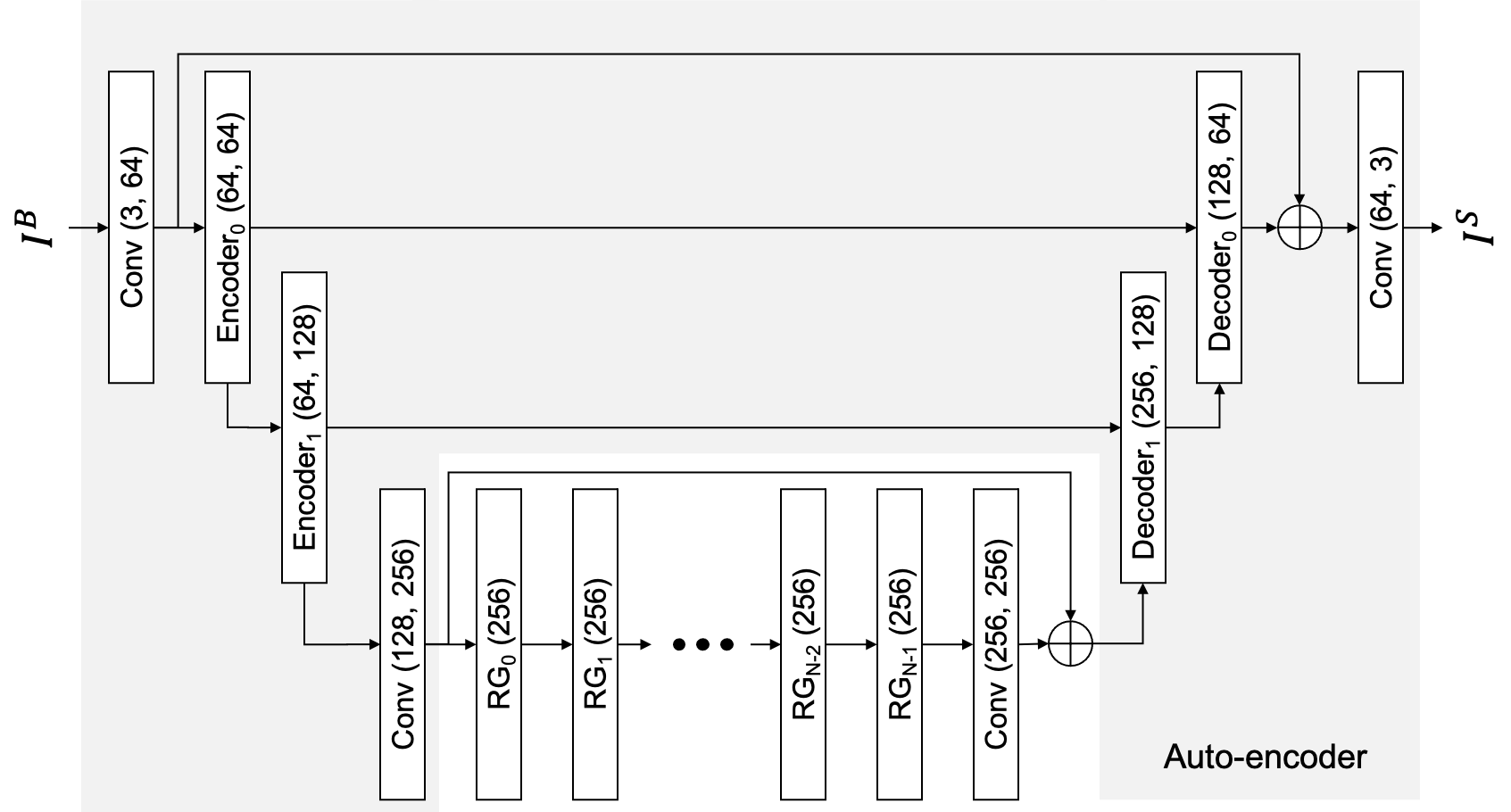}
    \\
    \figcspace
    \caption{
        \textbf{CAPP\_OB team (Track 2)}. Wide Receptive Field and Channel Attention Network
    }
    \label{fig:t2_capp}
    \figspace
\end{figure}

CAPP\_OB team proposed a wide receptive field and channel attention network~(WRCAN), an encoder-decoder architecture similar to UNet~\cite{ronneberger2015u}.
Dilated convolutions are used to increase the receptive field and the channel attention~\cite{Zhang_2018_ECCV_rcan} considers the relation between the feature channels.
They further optimize the model using with auto-encoder loss~\cite{Kwak_2019_CVPR_Workshops} to handle the JPEG compression artifacts.
The overall architecture is shown in Figure~\ref{fig:t2_capp}.
Please refer to \cite{Lee_2021_CVPR_Workshops_Wide} for more details.

\subsection{DuLang}

\begin{figure}[!h]
    \centering
    \includegraphics[width=\linewidth]{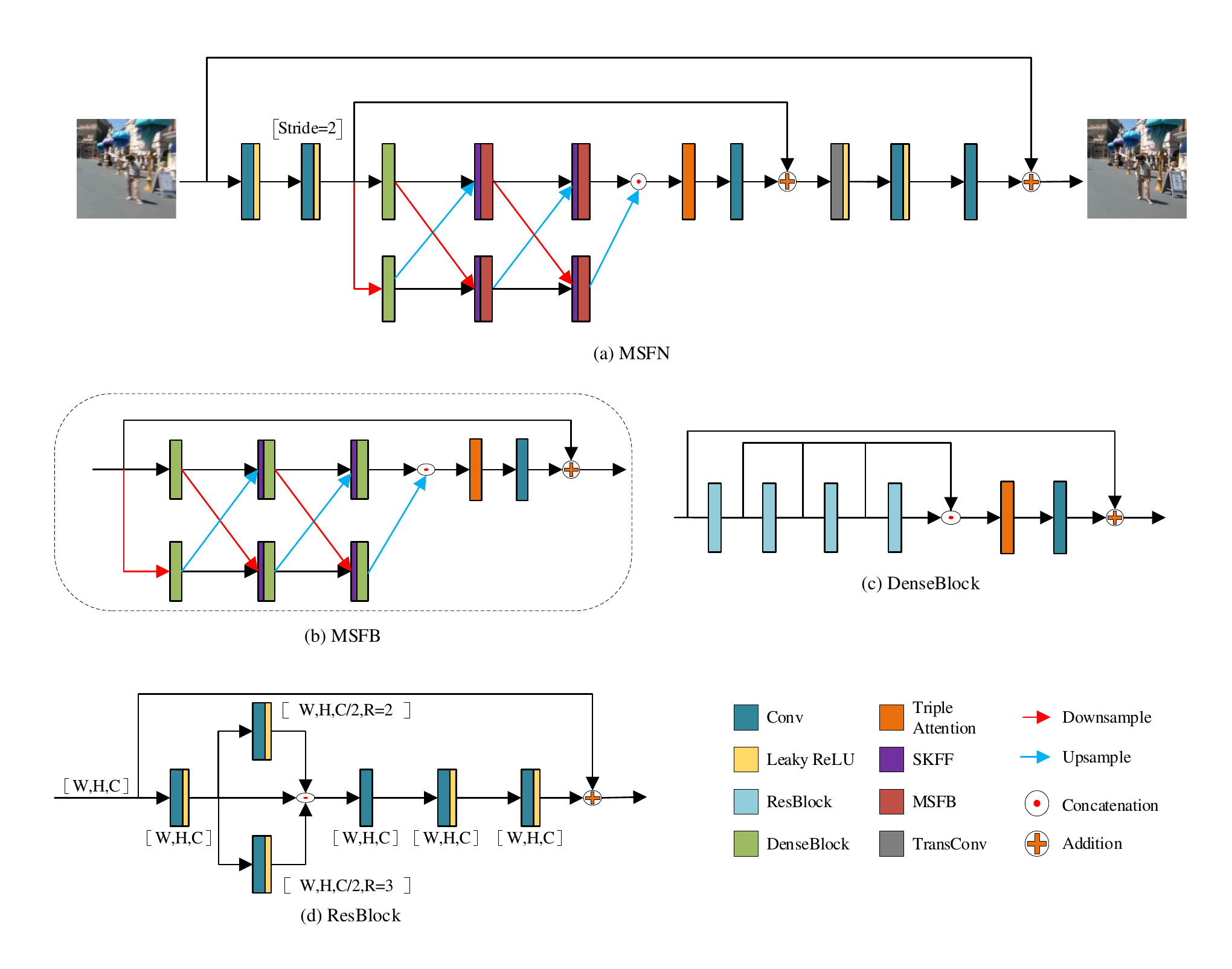}
    \\
    \figcspace
    \caption{
        \textbf{DuLang team (Track 2)}. Multi-Scale Fusion Net
        }
    \label{fig:t2_dulang}
    \figspace
\end{figure}

DuLang team proposes a multi-scale Fusion Net~(MSFN) based on AFN~\cite{Xu_2020_CVPR_Workshops} and MIRNet~\cite{zamir2020learning} to restore blurry images with JPEG artifacts.
To expand the receptive field, dilated convolutions are added to ResBlock.
Triple attention computes the attention weights by capturing cross-dimension interaction using a three-branch structure~\cite{Misra_2021_WACV}.
To train the proposed model, L1 loss, L1 loss between the Lapalacian images, and the L1 loss between the Sobel-filtered images are used.
The overall architecture is shown in Figure~\ref{fig:t2_dulang}.

\subsection{GiantPandaCV}

\begin{figure}[h]
    \centering
    \includegraphics[width=\linewidth]{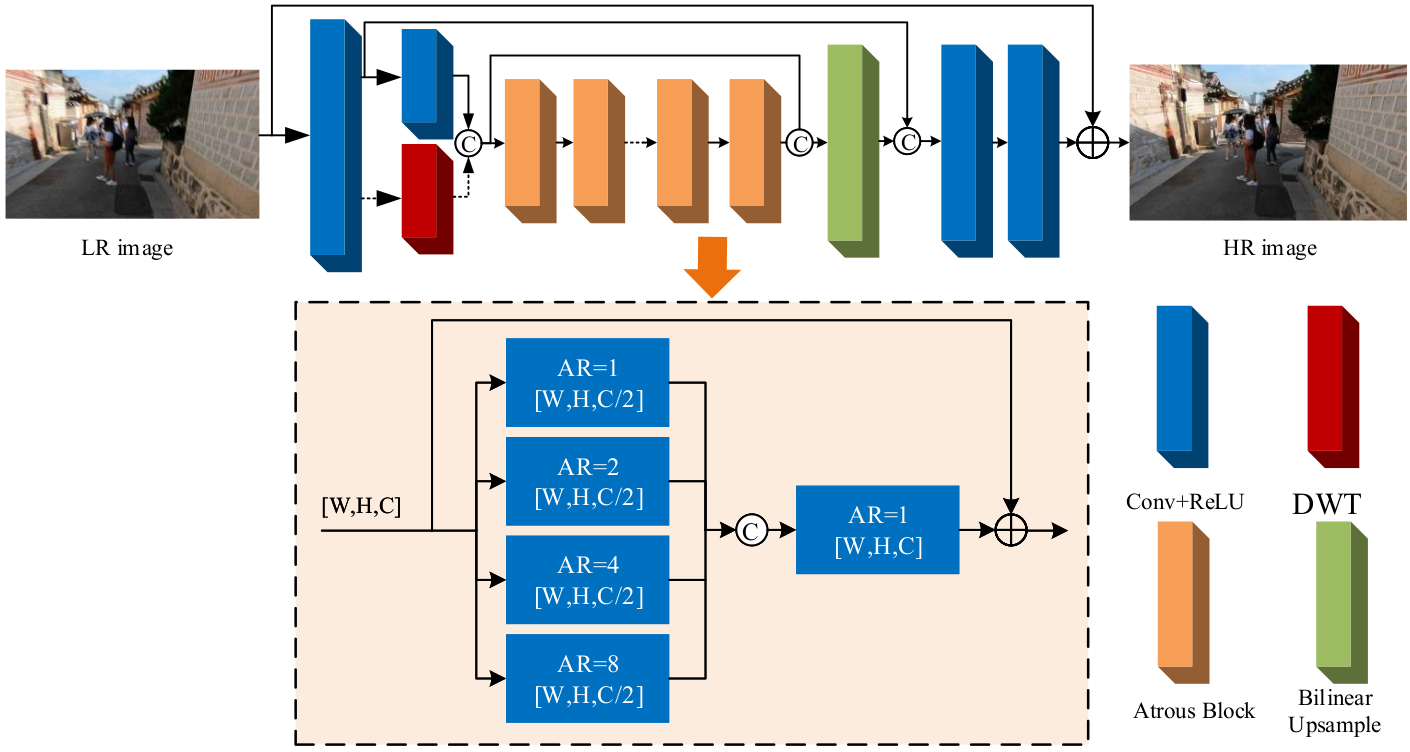}
    \\
    \figcspace
    \caption{
        \textbf{GiantPandaCV team (Track 2)}. A Simple Dilated Encoder-Decoder Network
    }
    \label{fig:t2_giantpandacv}
    \figspace
\end{figure}

GiantPandaCV team proposed a simple encoder-decoder structure model.
Different from U-Net, the receptive field is enlarged by dilated convolution layers.
They used SSIM and Charbonnier loss function to train the proposed model.
The overall architecture is shown in Figure~\ref{fig:t2_giantpandacv}.

\subsection{Maradona}

Maradona team used a multi-scale residual network model~\cite{Nah_2017_CVPR} by extending the model depth to 182 layers in their participation in track 2.
$\times8$ self-ensemble was used at test time.

\subsection{LAB FUD}

\begin{figure}[h]
    \centering
    \includegraphics[width=\linewidth]{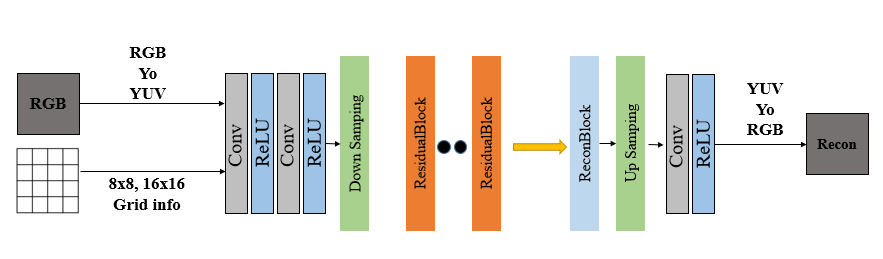}
    \\
    \figcspace
    \caption{
        \textbf{LAB FUD team (Track 2)}. yuv-grid-net
    }
    \label{fig:t2_labfud}
    \figspace
\end{figure}

LAB FUD team proposed a yuv-grid-net.
The model convers the input sRGB image to YUV colorspace and concatenates a grid map in $8\times8$ and $16\times16$ size.
Processed by a residual network, the output of the model is converted from the YUV colorspace to sRGB.
The overall architecture is shown in Figure~\ref{fig:t2_labfud}.

\subsection{SYJ}

\begin{figure}[h]
    \centering
    \includegraphics[width=\linewidth]{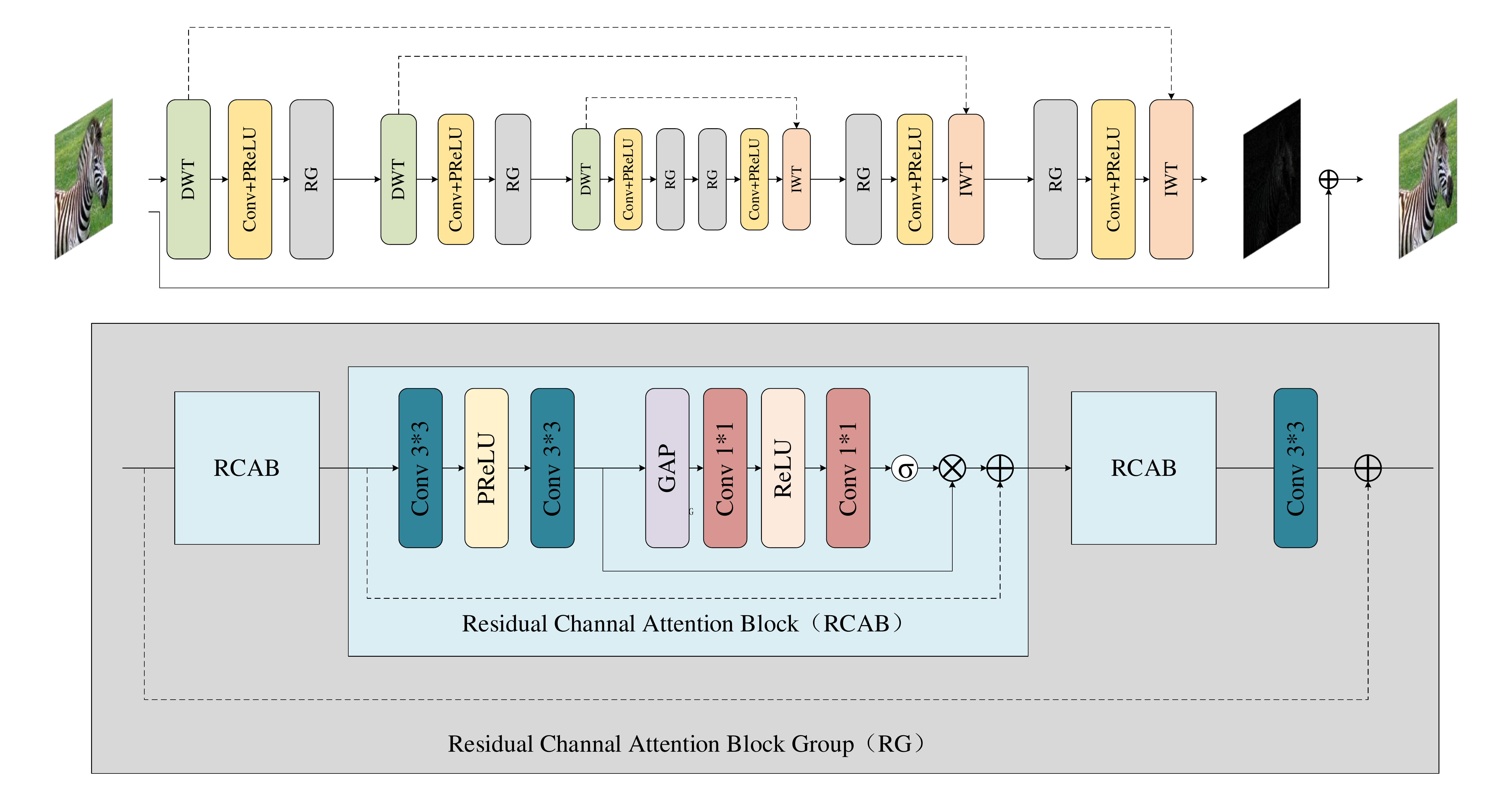}
    \\
    \figcspace
    \caption{
        \textbf{SYJ team (Track 2)}. Multi-level Wavelet-ResNet in Residual Learning
    }
    \label{fig:t2_syj}
    \figspace
\end{figure}

SYJ team proposed a Multi-level Wavelet-ResNet.
The proposed method performs discrete wavelet transforms in the neural network.
Residual group modules~\cite{Zhang_2018_ECCV_rcan} are used in the model with intermediate residual connections.
The overall architecture is shown in Figure~\ref{fig:t2_syj}.

\subsection{Dseny}

\begin{figure}[h]
    \centering
    \includegraphics[width=\linewidth]{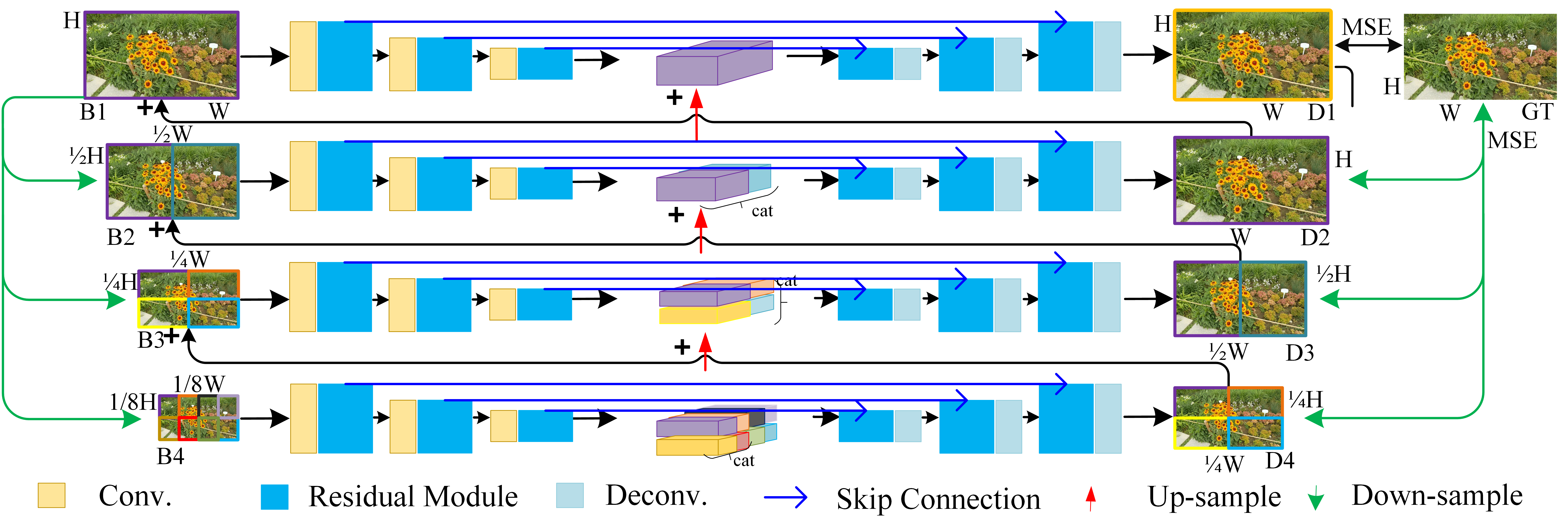}
    \\
    \figcspace
    \caption{
        \textbf{Dseny team (Track 2)}. Multi-scale and Multi-patch Network
    }
    \label{fig:t2_dseny}
    \figspace
\end{figure}

Dseny team presented a multi-scale and multi-patch network to deblur images in real-time.
They combined SRN~\cite{Tao_2018_CVPR} and DMPHN~\cite{Zhang_2019_CVPR_DMPHN} to build their model architecture.
The overall architecture is shown in Figure~\ref{fig:t2_dseny}.

\subsection{IPCV\_IITM}

\begin{figure}[h]
    \centering
    \includegraphics[width=\linewidth]{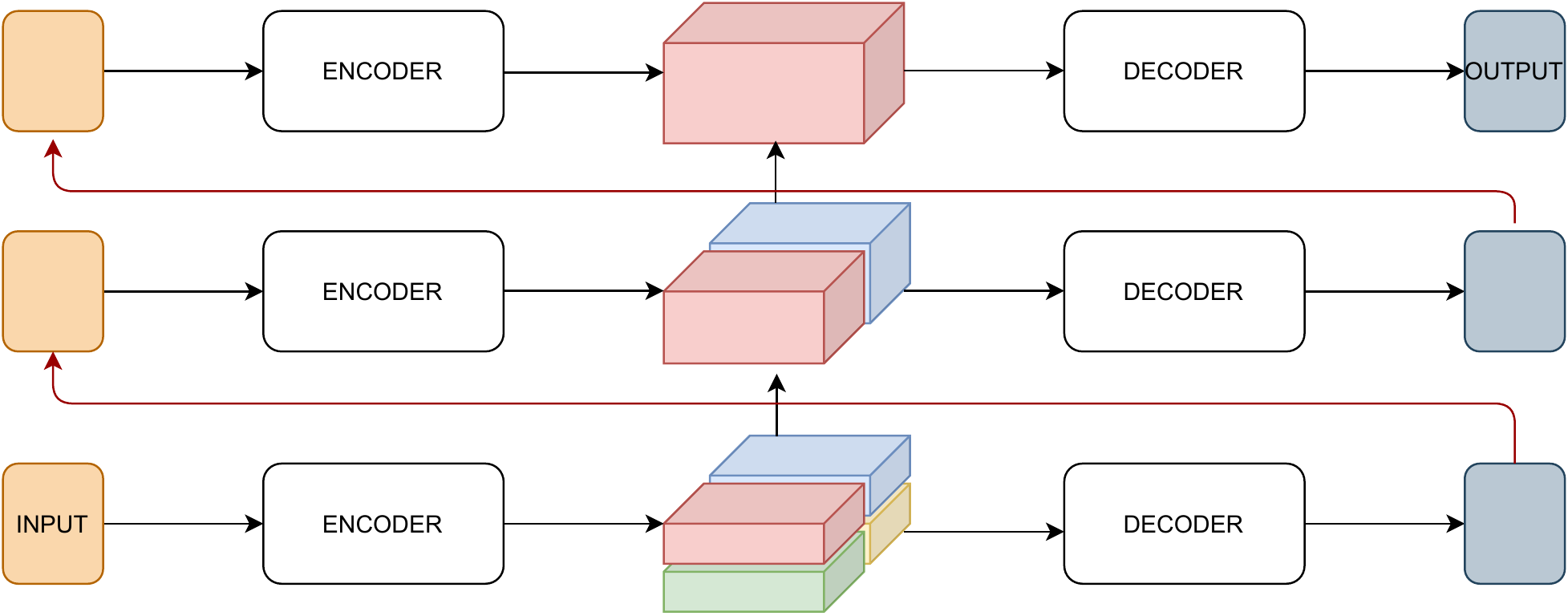}
    \\
    \figcspace
    \caption{
        \textbf{IPCV\_IITM team (Track 2)}. Hierarchical Encoder-Decoder with Multi-scale Convolution
    }
    \label{fig:t2_ipcv}
    \figspace
\end{figure}

IPCV\_IITM team used multi-scale context block~\cite{yu2015multi} in multi-patch hierarchical~\cite{Zhang_2019_CVPR_DMPHN,Suin_2020_CVPR} architecture.
The overall architecture is shown in Figure~\ref{fig:t2_ipcv}.

\subsection{Blur Attack}

\begin{figure}[h]
    \centering
    \includegraphics[width=\linewidth]{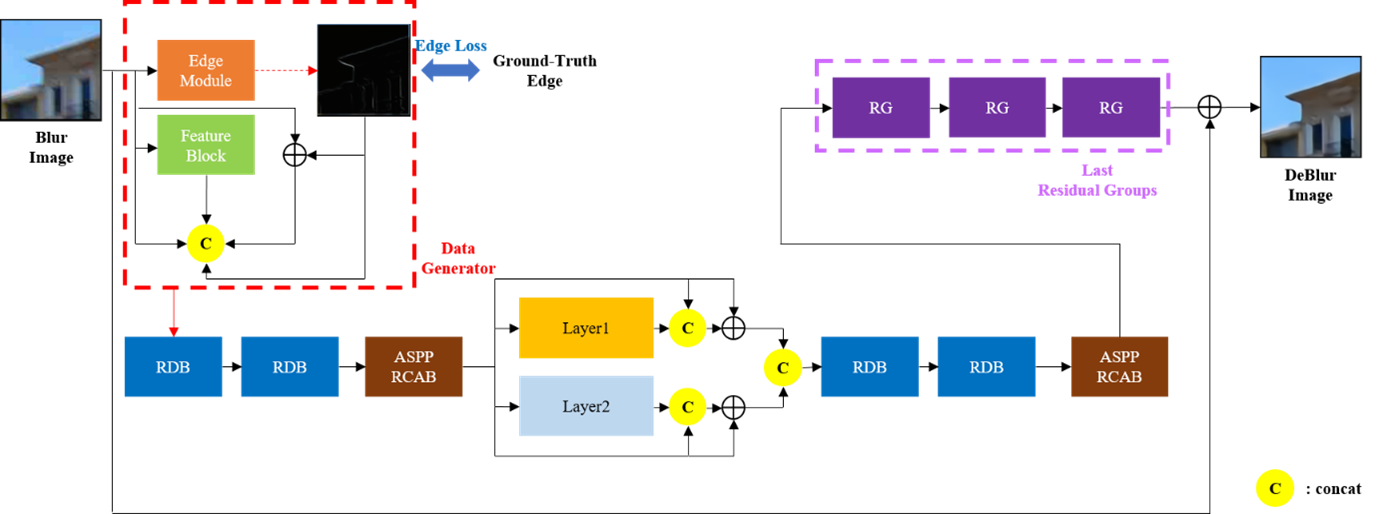}
    \\
    \figcspace
    \caption{
        \textbf{Blur Attack team (Track 2)}. EACD: Deblurring Network Using Edge Module, ASPP Channel Attention and Dual Network
    }
    \label{fig:t2_blur}
    \figspace
\end{figure}

Blur Attack team proposed a model named EACD.
The model extracts edge in addition to the convolutional feature.
The features are processed by residual dense blocks~\cite{Zhang_2018_CVPR_rdn} and residual groups~\cite{Zhang_2018_ECCV_rcan}.
The overall architecture is shown in Figure~\ref{fig:t2_blur}.

\section*{Acknowledgments}

We thank the NTIRE 2021 sponsors: HUAWEI Technologies Co. Ltd., Wright Brothers Institute, Facebook Reality Labs, MediaTek Inc., OPPO Mobile Corp., Ltd. and ETH Zurich (Computer Vision Lab).

\newpage
\appendix
\section{Teams and affiliations}
\label{sec:appendix}
\subsection*{NTIRE 2021 team}
\noindent\textit{\textbf{Title: }}NTIRE 2021 Challenge on Image Deblurring\\
\noindent\textit{\textbf{Members: }}\textit{Seungjun Nah$^1$ (seungjun.nah@gmail.com)}, Sanghyun Son$^1$, Suyoung Lee$^1$, Radu Timofte$^2$,  Kyoung Mu Lee$^1$\\
\noindent\textit{\textbf{Affiliations: }}\\
$^1$ Department of ECE, ASRI, SNU, Korea\\
$^2$ Computer Vision Lab, ETH Zurich, Switzerland\\

\subsection*{VIDAR}
\noindent\textit{\textbf{Title: }}EDPN: Enhanced Deep Pyramid Network for Blurry Image Restoration\\
\noindent\textit{\textbf{Members: }}\textit{Zhiwei Xiong (zwxiong@ustc.edu.cn)}, Ruikang Xu, Zeyu Xiao, Jie Huang, Yueyi Zhang\\
\noindent\textit{\textbf{Affiliations: }}\\
University of Science and Technology of China, China\\

\subsection*{The Fat, The Thin and The Strong}
\noindent\textit{\textbf{Title: }}HINet: Half Instance Normalization Network for Image Restoration\\
\noindent\textit{\textbf{Members: }}\textit{Liangyu Chen$^1$ (chenliangyu@megvii.com)}, Jie Zhang$^2$, Xin Lu$^1$, Xiaojie Chu$^3$, Chengpeng Chen$^1$\\
\noindent\textit{\textbf{Affiliations: }}\\
$^1$ Megvii\\
$^2$ Fudan University, China\\
$^3$ Peking University, China\\

\subsection*{netai}
\noindent\textit{\textbf{Title: }}Enhanced Multi-task Network for Blur Image Super-Resolution\\
\noindent\textit{\textbf{Members: }}\textit{Si Xi (xisi789@126.com)}, Jia Wei\\
\noindent\textit{\textbf{Affiliations: }}\\
Netease Games AI Lab\\

\subsection*{NJUST-IMAG}
\noindent\textit{\textbf{Title: }}Learning A Cascaded Non-Local Residual Network for Super-resolving Blurry Images\\
\noindent\textit{\textbf{Members: }}\textit{Haoran Bai$^1$ (baihaoran@njust.edu.cn)}, Songsheng Cheng$^1$, Hao Wei$^1$, Long Sun$^2$, Jinhui Tang$^1$, Jinshan Pan$^1$\\
\noindent\textit{\textbf{Affiliations: }}\\
$^1$ Nanjing University of Science and Technology, China\\
$^2$ Guilin University of Electronic Technology, China\\

\subsection*{CAPP\_OB}
\noindent\textit{\textbf{Title: }}Wide Receptive Field and Channel Attention Network for Deblurring of JPEG Compressed Image\\
\noindent\textit{\textbf{Members: }}\textit{Donghyeon Lee$^1$ (donghyeon1223@gmail.com)}, Chulhee Lee$^1$, Taesung Kim$^2$\\
\noindent\textit{\textbf{Affiliations: }}\\
$^1$ Samsung Electronics Co., Ltd.\\
$^2$ Sunmoon University, Asan, Korea\\

\subsection*{SRC-B}
\noindent\textit{\textbf{Title: }}MRNet: Multi-Refinement Network for Image Deblurring\\
\noindent\textit{\textbf{Members: }}\textit{Xiaobing Wang (x0106.wang@samsung.com)}, Dafeng Zhangr\\
\noindent\textit{\textbf{Affiliations: }}\\
Samsung Research China - Beijing\\

\subsection*{BAIDU}
\noindent\textit{\textbf{Title: }}Joint Super-Resolution and Deblurring Using Dual-Model Ensemble\\
\noindent\textit{\textbf{Members: }}\textit{Zhihong Pan$^1$ (zhihongpan@baidu.com)}, Tianwei Lin$^2$, Wenhao Wu$^2$, Dongliang He$^2$, Baopu Li$^1$, Boyun Li$^2$, Teng Xi$^2$, Gang Zhang$^2$, Jingtuo Liu$^2$, Junyu Han$^2$, Errui Ding$^2$\\
\noindent\textit{\textbf{Affiliations: }}\\
$^1$ Baidu Research USA\\
$^2$ Department of Computer Vision Technology, Baidu, Inc.\\

\subsection*{MMM}
\noindent\textit{\textbf{Title: }}M3Net: Multi-stage, Multi-patch and Multi-resolution for image deblurring\\
\noindent\textit{\textbf{Members: }}\textit{Jie Zhang$^1$ (j\_zhang19@fudan.edu.cn)}, Liangyu Chen$^2$, Chengpeng Chen$^2$, Xin Lu$^2$, Xiaojie Chu$^3$\\
\noindent\textit{\textbf{Affiliations: }}\\
$^1$ Fudan University, China\\
$^2$ Megvii\\
$^3$ Peking University, China\\

\subsection*{Imagination}
\noindent\textit{\textbf{Title: }}Pyramid Deformable Convolution\\
\noindent\textit{\textbf{Members: }}\textit{Guangping Tao$^1$ (tgpin@smail.nju.edu.cn)}, Wenqing Chu$^2$, Yun Cao$^2$, Donghao Luo$^2$, Ying Tai$^2$, Tong Lu$^1$, Chengjie Wang$^2$, Jilin Li$^2$, Feiyue Huang$^2$\\
\noindent\textit{\textbf{Affiliations: }}\\
$^1$ Nanjing University\\
$^2$ Tencent\\

\subsection*{Noah\_CVlab}
\noindent\textit{\textbf{Title: }}Pre-Trained Image Processing Transformer\\
\noindent\textit{\textbf{Members: }}\textit{Hanting Chen (chenhanting@huawei.com)}, Shuaijun Chen, Tianyu Guo, Yunhe Wang\\
\noindent\textit{\textbf{Affiliations: }}\\
Noah's Ark Lab, Huawei Technologies Co., Ltd.\\

\subsection*{TeamInception}
\noindent\textit{\textbf{Title: }}Multi-Stage Progressive Image Restoration\\
\noindent\textit{\textbf{Members: }}\textit{Syed Waqas Zamir (waqas.zamir@inceptioniai.org)}, Aditya Arora, Salman Khan, Munawar Hayat, Fahad Shahbaz Kahn, Ling Shao\\
\noindent\textit{\textbf{Affiliations: }}\\
Inception Institute of Artificial Intelligence\\

\subsection*{ZOCS\_Team}
\noindent\textit{\textbf{Title: }}PyVTRDN - Equip RDN with Pyramid token based transformer\\
\noindent\textit{\textbf{Members: }}\textit{Yushen Zuo (zys19@mails.tsinghua.edu.cn)}, Yimin Ou, Yuanjun Chai, Lei Shi\\
\noindent\textit{\textbf{Affiliations: }}\\
Tsinghua University, Beijing, China\\

\subsection*{Mier}
\noindent\textit{\textbf{Title: }}Big UNet for Image Restoration\\
\noindent\textit{\textbf{Members: }}\textit{Shuai Liu$^1$ (18601200232@163.com)}, Lei Lei$^2$, Chaoyu Feng$^2$\\
\noindent\textit{\textbf{Affiliations: }}\\
$^1$ North China University of Technology\\
$^2$ Xiaomi\\

\subsection*{INFINITY}
\noindent\textit{\textbf{Title: }}Image Deblurring with Enhanced Deep Residual Network\\
\noindent\textit{\textbf{Members: }}\textit{Kai Zeng (eezk@mail.scut.edu.cn)}, Yuying Yao, Xinran Liu\\
\noindent\textit{\textbf{Affiliations: }}\\
South China University of Technology, China\\

\subsection*{DuLang}
\noindent\textit{\textbf{Title: }}Multi-Scale Fusion Net for Single Image Deblurring\\
\noindent\textit{\textbf{Members: }}\textit{Zhizhou Zhang (220201874@seu.edu.cn)}, Huacheng Huang\\
\noindent\textit{\textbf{Affiliations: }}\\
Lab of Image Science and Technology, Southeast University, China\\

\subsection*{GiantPandaCV}
\noindent\textit{\textbf{Title: }}A Simple Dilated Encoder-Decoder Network for Image Restoration\\
\noindent\textit{\textbf{Members: }}\textit{Yunchen Zhang (cydiachen@cydiachen.tech)}, Mingchao Jiang, Wenbin Zou\\
\noindent\textit{\textbf{Affiliations: }}\\
$^1$ China Design Group Co., Ltd.\\
$^2$ JOYY AI GROUP\\
$^3$ Fujian Normal University, China\\

\subsection*{Maradona}
\noindent\textit{\textbf{Title: }}Multi-scale Network for Image Deblurring\\
\noindent\textit{\textbf{Members: }}\textit{Si Miao (miaosi2018@gmail.com)}\\
\noindent\textit{\textbf{Affiliations: }}\\
Shanghai Advanced Research Institute, Chinese Academy of Sciences\\

\subsection*{LAB FUD}
\noindent\textit{\textbf{Title: }}yuv-grid-net\\
\noindent\textit{\textbf{Members: }}\textit{Yangwoo Kim (tuzm24@gmail.com)}\\

\subsection*{SYJ}
\noindent\textit{\textbf{Title: }}Multi-level Wavelet-Resnet in Residual learning\\
\noindent\textit{\textbf{Members: }}\textit{Yuejin Sun (yjsun97@mail.ustc.edu.cn)}\\
\noindent\textit{\textbf{Affiliations: }}\\
University of Science and Technology of China, China\\

\subsection*{Dseny}
\noindent\textit{\textbf{Title: }}Multi-scale and Multi-patch Network for High-Definition Image Deblurring in Real-time\\
\noindent\textit{\textbf{Members: }}\textit{Senyou Deng$^1$ (dengsenyou@iie.ac.cn)}, Wenqi Ren$^1$, Xiaochun Cao$^1$, Tao Wang$^2$\\
\noindent\textit{\textbf{Affiliations: }}\\
$^1$ Institute of Information Engineering, Chinese Academy of Sciences, China\\
$^2$ Huawei Noah's Ark Lab\\

\subsection*{IPCV\_IITM}
\noindent\textit{\textbf{Title: }}Hierarchical Encoder-Decoder with Multi-scale Convolution for Image Deblurring\\
\noindent\textit{\textbf{Members: }}\textit{Maitreya Suin (maitreyasuin21@gmail.com)}, A. N. Rajagopalan\\
\noindent\textit{\textbf{Affiliations: }}\\
Indian Institute of Technology Madras, India\\

\subsection*{DMLAB}
\noindent\textit{\textbf{Title: }}Multi-scale Hierarchical Dense Residual Network for Image Deblurring\\
\noindent\textit{\textbf{Members: }}\textit{Vinh Van Duong (duongvinh@skku.edu)}, Thuc Huu Nguyen, Jonghoon Yim and Byeungwoo Jeon\\
\noindent\textit{\textbf{Affiliations: }}\\
Department of ECE, Sungkyunkwan University, Korea\\

\subsection*{RTQSA-Lab}
\noindent\textit{\textbf{Title: }}Enhanced Attention Network for Single Image Deblurring\\
\noindent\textit{\textbf{Members: }}\textit{Ru Li$^1$ (leeru564636@163.com)}, Junwei Xie$^2$\\
\noindent\textit{\textbf{Affiliations: }}\\
$^1$ Fuzhou University, China\\
$^2$ Imperial Vision Co. Ltd.\\

\subsection*{Yonsei-MCML}
\noindent\textit{\textbf{Title: }}Edge Attention Network for Image Deblurring and Super-Resolution\\
\noindent\textit{\textbf{Members: }}\textit{Jong-Wook Han (johnhan00@yonsei.ac.kr)}, Jun-Ho Choi, Jun-Hyuk Kim, Jong-Seok Lee\\
\noindent\textit{\textbf{Affiliations: }}\\
School of Integrated Technology, Yonsei University\\

\subsection*{SCUT-ZS}
\noindent\textit{\textbf{Title: }}EDSR for NTIRE 2021 Image Deblurring Challenge\\
\noindent\textit{\textbf{Members: }}\textit{Jiaxin Zhang (msjxzhang@mail.scut.edu.cn)}, Fan Peng\\
\noindent\textit{\textbf{Affiliations: }}\\
South China University of Technology, China\\


\subsection*{Expasoft team}
\noindent\textit{\textbf{Title: }}BowNet\\
\noindent\textit{\textbf{Members: }}\textit{David Svitov (d.svitov@expasoft.tech)}, Dmitry Pakulich\\
\noindent\textit{\textbf{Affiliations: }}\\
$^1$ Expasoft LLC\\
$^2$ Institute of Automation and Electrometry of the SB RAS\\

\subsection*{Blur Attack}
\noindent\textit{\textbf{Title: }}EACD : Deblurring Network Using Edge Module, ASPP Channel Attention and Dual Network\\
\noindent\textit{\textbf{Members: }}\textit{Jaeyeob Kim (athurk94111@gmail.com)}, Jechang Jeong\\
\noindent\textit{\textbf{Affiliations: }}\\
Image Communication \& Signal Processing Laboratory, Hanyang University, Korea\\

\newpage
{\small
\bibliographystyle{ieee_fullname}
\bibliography{egbib}

\begin{thebibliography}{100}\itemsep=-1pt

\bibitem{abuolaim2021ntire}
Abdullah Abuolaim, Radu Timofte, Michael~S Brown, et~al.
\newblock {NTIRE 2021} challenge for defocus deblurring using dual-pixel
  images: Methods and results.
\newblock In {\em CVPR Workshops}, 2021.

\bibitem{Agustsson_2017_CVPR_Workshops_div2k}
Eirikur Agustsson and Radu Timofte.
\newblock {NTIRE} 2017 challenge on single image super-resolution: Dataset and
  study.
\newblock In {\em CVPR Workshops}, 2017.

\bibitem{ancuti2021ntire}
Codruta~O Ancuti, Cosmin Ancuti, Florin-Alexandru Vasluianu, Radu Timofte,
  et~al.
\newblock {NTIRE 2021} nonhomogeneous dehazing challenge report.
\newblock In {\em CVPR Workshops}, 2021.

\bibitem{anwar2020densely}
Saeed Anwar and Nick Barnes.
\newblock Densely residual laplacian super-resolution.
\newblock {\em IEEE TPAMI}, 2020.

\bibitem{bai2021cascaded}
Haoran Bai, Songsheng Cheng, Jinhui Tang, and Jinshan Pan.
\newblock Learning a cascaded non-local residual network for super-resolving
  blurry images.
\newblock In {\em CVPR Workshops}, 2021.

\bibitem{bhat2021ntire}
Goutam Bhat, Martin Danelljan, Radu Timofte, et~al.
\newblock {NTIRE 2021} challenge on burst super-resolution: Methods and
  results.
\newblock In {\em CVPR Workshops}, 2021.

\bibitem{Brehm_2020_CVPR_Workshops}
Stephan Brehm, Sebastian Scherer, and Rainer Lienhart.
\newblock High-resolution dual-stage multi-level feature aggregation for single
  image and video deblurring.
\newblock In {\em CVPR Workshops}, 2020.

\bibitem{Cai_2019_ICCV_toward_real}
Jianrui Cai, Hui Zeng, Hongwei Yong, Zisheng Cao, and Lei Zhang.
\newblock Toward real-world single image super-resolution: A new benchmark and
  a new model.
\newblock In {\em ICCV}, 2019.

\bibitem{chang2013reducing}
Huibin Chang, Michael~K Ng, and Tieyong Zeng.
\newblock Reducing artifacts in jpeg decompression via a learned dictionary.
\newblock {\em IEEE Transactions on Signal Processing}, 62(3):718--728, 2013.

\bibitem{chen2020pre}
Hanting Chen, Yunhe Wang, Tianyu Guo, Chang Xu, Yiping Deng, Zhenhua Liu, Siwei
  Ma, Chunjing Xu, Chao Xu, and Wen Gao.
\newblock Pre-trained image processing transformer.
\newblock {\em arXiv preprint arXiv:2012.00364}, 2020.

\bibitem{chen2021hinet}
Liangyu Chen, Xin Lu, Jie Zhang, Xiaojie Chu, and Chengpeng Chen.
\newblock {HIN}et: Half instance normalization network for image restoration.
\newblock In {\em CVPR Workshops}, 2021.

\bibitem{cho2009fast}
Sunghyun Cho and Seungyong Lee.
\newblock Fast motion deblurring.
\newblock In {\em ACM SIGGRAPH Asia}, 2009.

\bibitem{Dai_2019_CVPR_second_order}
Tao Dai, Jianrui Cai, Yongbing Zhang, Shu-Tao Xia, and Lei Zhang.
\newblock Second-order attention network for single image super-resolution.
\newblock In {\em CVPR}, 2019.

\bibitem{deng2009imagenet}
Jia Deng, Wei Dong, Richard Socher, Li-Jia Li, Kai Li, and Li Fei-Fei.
\newblock Image{N}et: A large-scale hierarchical image database.
\newblock In {\em CVPR}, 2009.

\bibitem{dong2015compression}
Chao Dong, Yubin Deng, Chen~Change Loy, and Xiaoou Tang.
\newblock Compression artifacts reduction by a deep convolutional network.
\newblock In {\em ICCV}, 2015.

\bibitem{dong2014learning_srcnn}
Chao Dong, Chen~Change Loy, Kaiming He, and Xiaoou Tang.
\newblock Learning a deep convolutional network for image super-resolution.
\newblock In {\em ECCV}, 2014.

\bibitem{dong2016accelerating}
Chao Dong, Chen~Change Loy, and Xiaoou Tang.
\newblock Accelerating the super-resolution convolutional neural network.
\newblock In {\em ECCV}, 2016.

\bibitem{elhelou2021ntire}
Majed El~Helou, Ruofan Zhou, Sabine S\"usstrunk, Radu Timofte, et~al.
\newblock {NTIRE 2021} depth guided image relighting challenge.
\newblock In {\em CVPR Workshops}, 2021.

\bibitem{Fu_2019_ICCV}
Xueyang Fu, Zheng-Jun Zha, Feng Wu, Xinghao Ding, and John Paisley.
\newblock {JPEG} artifacts reduction via deep convolutional sparse coding.
\newblock In {\em ICCV}, 2019.

\bibitem{Gao_2019_CVPR}
Hongyun Gao, Xin Tao, Xiaoyong Shen, and Jiaya Jia.
\newblock Dynamic scene deblurring with parameter selective sharing and nested
  skip connections.
\newblock In {\em CVPR}, 2019.

\bibitem{gu2021ntire}
Jinjin Gu, Haoming Cai, Chao Dong, Jimmy~S. Ren, Yu Qiao, Shuhang Gu, Radu
  Timofte, et~al.
\newblock {NTIRE 2021} challenge on perceptual image quality assessment.
\newblock In {\em CVPR Workshops}, 2021.

\bibitem{Gu_2019_CVPR_ikc}
Jinjin Gu, Hannan Lu, Wangmeng Zuo, and Chao Dong.
\newblock Blind super-resolution with iterative kernel correction.
\newblock In {\em CVPR}, 2019.

\bibitem{guo2016building}
Jun Guo and Hongyang Chao.
\newblock Building dual-domain representations for compression artifacts
  reduction.
\newblock In {\em ECCV}, 2016.

\bibitem{guo2017one}
Jun Guo and Hongyang Chao.
\newblock One-to-many network for visually pleasing compression artifacts
  reduction.
\newblock In {\em CVPR}, 2017.

\bibitem{Guo_2020_CVPR_closed_loop}
Yong Guo, Jian Chen, Jingdong Wang, Qi Chen, Jiezhang Cao, Zeshuai Deng, Yanwu
  Xu, and Mingkui Tan.
\newblock Closed-loop matters: Dual regression networks for single image
  super-resolution.
\newblock In {\em CVPR}, 2020.

\bibitem{Haris_2018_CVPR}
Muhammad Haris, Gregory Shakhnarovich, and Norimichi Ukita.
\newblock Deep back-projection networks for super-resolution.
\newblock In {\em CVPR}, 2018.

\bibitem{He_2016_ECCV}
Kaiming He, Xiangyu Zhang, Shaoqing Ren, and Jian Sun.
\newblock Identity mappings in deep residual networks.
\newblock In {\em ECCV}, 2016.

\bibitem{He_2019_CVPR_oisr}
Xiangyu He, Zitao Mo, Peisong Wang, Yang Liu, Mingyuan Yang, and Jian Cheng.
\newblock {ODE}-inspired network design for single image super-resolution.
\newblock In {\em CVPR}, 2019.

\bibitem{Hu_2019_CVPR_metasr}
Xuecai Hu, Haoyuan Mu, Xiangyu Zhang, Zilei Wang, Tieniu Tan, and Jian Sun.
\newblock Meta-{SR}: A magnification-arbitrary network for super-resolution.
\newblock In {\em CVPR}, 2019.

\bibitem{ioffe2015batch}
Sergey Ioffe and Christian Szegedy.
\newblock Batch normalization: Accelerating deep network training by reducing
  internal covariate shift.
\newblock {\em arXiv preprint arXiv:1502.03167}, 2015.

\bibitem{kim2016accurate_vdsr}
Jiwon Kim, Jung~Kwon Lee, and Kyoung~Mu Lee.
\newblock Accurate image super-resolution using very deep convolutional
  networks.
\newblock In {\em CVPR}, 2016.

\bibitem{Kim_2013_ICCV}
Tae~Hyun Kim, Byeongjoo Ahn, and Kyoung~Mu Lee.
\newblock Dynamic scene deblurring.
\newblock In {\em ICCV}, 2013.

\bibitem{Kim_2014_CVPR}
Tae~Hyun Kim and Kyoung~Mu Lee.
\newblock Segmentation-free dynamic scene deblurring.
\newblock In {\em CVPR}, 2014.

\bibitem{Kupyn_2018_CVPR}
Orest Kupyn, Volodymyr Budzan, Mykola Mykhailych, Dmytro Mishkin, and Jiří
  Matas.
\newblock Deblur{GAN}: Blind motion deblurring using conditional adversarial
  networks.
\newblock In {\em CVPR}, 2018.

\bibitem{Kupyn_2019_ICCV}
Orest Kupyn, Tetiana Martyniuk, Junru Wu, and Zhangyang Wang.
\newblock Deblur{GAN}-v2: Deblurring (orders-of-magnitude) faster and better.
\newblock In {\em ICCV}, 2019.

\bibitem{Kwak_2019_CVPR_Workshops}
Junhyung Kwak and Donghee Son.
\newblock Fractal residual network and solutions for real super-resolution.
\newblock In {\em CVPR Workshops}, 2019.

\bibitem{Lai_2017_CVPR}
Wei-Sheng Lai, Jia-Bin Huang, Narendra Ahuja, and Ming-Hsuan Yang.
\newblock Deep laplacian pyramid networks for fast and accurate
  super-resolution.
\newblock In {\em CVPR}, 2017.

\bibitem{Ledig_2017_CVPR}
Christian Ledig, Lucas Theis, Ferenc Huszar, Jose Caballero, Andrew Cunningham,
  Alejandro Acosta, Andrew Aitken, Alykhan Tejani, Johannes Totz, Zehan Wang,
  and Wenzhe Shi.
\newblock Photo-realistic single image super-resolution using a generative
  adversarial network.
\newblock In {\em CVPR}, 2017.

\bibitem{Lee_2021_CVPR_Workshops_Wide}
Donghyeon Lee, Chulhee Lee, and Taesung Kim.
\newblock Wide receptive field and channel attention network for jpeg
  compressed image deblurring.
\newblock In {\em CVPR Workshops}, 2021.

\bibitem{levin2009understanding}
Anat Levin, Yair Weiss, Fredo Durand, and William~T Freeman.
\newblock Understanding and evaluating blind deconvolution algorithms.
\newblock In {\em CVPR}, 2009.

\bibitem{li2020mdcn}
Juncheng Li, Faming Fang, Jiaqian Li, Kangfu Mei, and Guixu Zhang.
\newblock {MDCN}: Multi-scale dense cross network for image super-resolution.
\newblock {\em IEEE TCSVT}, 2020.

\bibitem{li2014contrast}
Yu Li, Fangfang Guo, Robby~T Tan, and Michael~S Brown.
\newblock A contrast enhancement framework with jpeg artifacts suppression.
\newblock In {\em ECCV}, 2014.

\bibitem{Li_2019_CVPR_feedback}
Zhen Li, Jinglei Yang, Zheng Liu, Xiaomin Yang, Gwanggil Jeon, and Wei Wu.
\newblock Feedback network for image super-resolution.
\newblock In {\em CVPR}, 2019.

\bibitem{liew2004blocking}
AW-C Liew and Hong Yan.
\newblock Blocking artifacts suppression in block-coded images using
  overcomplete wavelet representation.
\newblock {\em IEEE TCSVT}, 14(4):450--461, 2004.

\bibitem{Lim_2017_CVPR_Workshops_edsr}
Bee Lim, Sanghyun Son, Heewon Kim, Seungjun Nah, and Kyoung~Mu Lee.
\newblock Enhanced deep residual networks for single image super-resolution.
\newblock In {\em CVPR Workshops}, 2017.

\bibitem{list2003adaptive}
Peter List, Anthony Joch, Jani Lainema, Gisle Bjontegaard, and Marta
  Karczewicz.
\newblock Adaptive deblocking filter.
\newblock {\em IEEE TCSVT}, 13(7):614--619, 2003.

\bibitem{liu2021ntire}
Jerrick Liu, Oliver Nina, Radu Timofte, et~al.
\newblock {NTIRE 2021} multi-modal aerial view object classification challenge.
\newblock In {\em CVPR Workshops}, 2021.

\bibitem{liu2020residual}
Jie Liu, Jie Tang, and Gangshan Wu.
\newblock Residual feature distillation network for lightweight image
  super-resolution.
\newblock {\em arXiv preprint arXiv:2009.11551}, 2020.

\bibitem{Liu_2020_CVPR_self_calibrated}
Jiang-Jiang Liu, Qibin Hou, Ming-Ming Cheng, Changhu Wang, and Jiashi Feng.
\newblock Improving convolutional networks with self-calibrated convolutions.
\newblock In {\em CVPR}, 2020.

\bibitem{Liu_2018_CVPR_Workshops}
Pengju Liu, Hongzhi Zhang, Kai Zhang, Liang Lin, and Wangmeng Zuo.
\newblock Multi-level wavelet-cnn for image restoration.
\newblock In {\em CVPR Workshops}, 2018.

\bibitem{Liu_2020_CVPR_Workshops}
Shuai Liu, Chenghua Li, Nan Nan, Ziyao Zong, and Ruixia Song.
\newblock {MMDM}: Multi-frame and multi-scale for image demoireing.
\newblock In {\em CVPR Workshops}, 2020.

\bibitem{liu2016random}
Xianming Liu, Gene Cheung, Xiaolin Wu, and Debin Zhao.
\newblock Random walk graph laplacian-based smoothness prior for soft decoding
  of jpeg images.
\newblock {\em IEEE TIP}, 26(2):509--524, 2016.

\bibitem{liu2015data}
Xianming Liu, Xiaolin Wu, Jiantao Zhou, and Debin Zhao.
\newblock Data-driven sparsity-based restoration of jpeg-compressed images in
  dual transform-pixel domain.
\newblock In {\em CVPR}, 2015.

\bibitem{Lu_2019_CVPR}
Boyu Lu, Jun-Cheng Chen, and Rama Chellappa.
\newblock Unsupervised domain-specific deblurring via disentangled
  representations.
\newblock In {\em CVPR}, 2019.

\bibitem{lugmayr2021ntire}
Andreas Lugmayr, Martin Danelljan, Radu Timofte, et~al.
\newblock {NTIRE 2021} learning the super-resolution space challenge.
\newblock In {\em CVPR Workshops}, 2021.

\bibitem{maas2013rectifier}
Andrew~L Maas, Awni~Y Hannun, and Andrew~Y Ng.
\newblock Rectifier nonlinearities improve neural network acoustic models.
\newblock In {\em ICML}, 2013.

\bibitem{Mei_2020_CVPR_cross_scale}
Yiqun Mei, Yuchen Fan, Yuqian Zhou, Lichao Huang, Thomas~S. Huang, and Honghui
  Shi.
\newblock Image super-resolution with cross-scale non-local attention and
  exhaustive self-exemplars mining.
\newblock In {\em CVPR}, 2020.

\bibitem{Misra_2021_WACV}
Diganta Misra, Trikay Nalamada, Ajay~Uppili Arasanipalai, and Qibin Hou.
\newblock Rotate to attend: Convolutional triplet attention module.
\newblock In {\em WACV}, 2021.

\bibitem{Nah_2019_CVPR_Workshops_REDS}
Seungjun Nah, Sungyong Baik, Seokil Hong, Gyeongsik Moon, Sanghyun Son, Radu
  Timofte, and Kyoung~Mu Lee.
\newblock {NTIRE} 2019 challenges on video deblurring and super-resolution:
  Dataset and study.
\newblock In {\em CVPR Workshops}, 2019.

\bibitem{Nah_2017_CVPR}
Seungjun Nah, Tae~Hyun Kim, and Kyoung~Mu Lee.
\newblock Deep multi-scale convolutional neural network for dynamic scene
  deblurring.
\newblock In {\em CVPR}, 2017.

\bibitem{Nah_2020_CVPR_Workshops_Deblur}
Seungjun Nah, Sanghyun Son, Radu Timofte, and Kyoung~Mu Lee.
\newblock {NTIRE} 2020 challenge on image and video deblurring.
\newblock In {\em CVPR Workshops}, 2020.

\bibitem{Nah_2019_CVPR_Workshops_Deblur}
Seungjun Nah, Radu Timofte, Sungyong Baik, Seokil Hong, Gyeongsik Moon,
  Sanghyun Son, and Kyoung~Mu Lee.
\newblock {NTIRE} 2019 challenge on video deblurring: Methods and results.
\newblock In {\em CVPR Workshops}, 2019.

\bibitem{Nah_2019_CVPR_Workshops_SR}
Seungjun Nah, Radu Timofte, Shuhang Gu, Sungyong Baik, Seokil Hong, Gyeongsik
  Moon, Sanghyun Son, and Kyoung~Mu Lee.
\newblock {NTIRE} 2019 challenge on video super-resolution: Methods and
  results.
\newblock In {\em CVPR Workshops}, 2019.

\bibitem{noroozi2017motion}
Mehdi Noroozi, Paramanand Chandramouli, and Paolo Favaro.
\newblock Motion deblurring in the wild.
\newblock In {\em GCPR}, 2017.

\bibitem{Park_2020_ECCV_MTRNN}
Dongwon Park, Dong~Un Kang, Jisoo Kim, and Se~Young Chun.
\newblock Multi-temporal recurrent neural networks for progressive non-uniform
  single image deblurring with incremental temporal training.
\newblock In {\em ECCV}, 2020.

\bibitem{perez2021ntire}
Eduardo P\'erez-Pellitero, Sibi Catley-Chandar, Ale\v{s} Leonardis, Radu
  Timofte, et~al.
\newblock {NTIRE 2021} challenge on high dynamic range imaging: Dataset,
  methods and results.
\newblock In {\em CVPR Workshops}, 2021.

\bibitem{rim2020real}
Jaesung Rim, Haeyun Lee, Jucheol Won, and Sunghyun Cho.
\newblock Real-world blur dataset for learning and benchmarking deblurring
  algorithms.
\newblock In {\em ECCV}, 2020.

\bibitem{ronneberger2015u}
Olaf Ronneberger, Philipp Fischer, and Thomas Brox.
\newblock U-net: Convolutional networks for biomedical image segmentation.
\newblock In {\em MICCAI}, 2015.

\bibitem{Shen_2019_ICCV_Human_aware}
Ziyi Shen, Wenguan Wang, Xiankai Lu, Jianbing Shen, Haibin Ling, Tingfa Xu, and
  Ling Shao.
\newblock Human-aware motion deblurring.
\newblock In {\em ICCV}, 2019.

\bibitem{Shi_2016_CVPR_espcn}
Wenzhe Shi, Jose Caballero, Ferenc Huszar, Johannes Totz, Andrew~P. Aitken, Rob
  Bishop, Daniel Rueckert, and Zehan Wang.
\newblock Real-time single image and video super-resolution using an efficient
  sub-pixel convolutional neural network.
\newblock In {\em CVPR}, 2016.

\bibitem{xiandwei2021pdan}
Xi Si, Wei Jia, and Zhang Weidong.
\newblock Pixel-guided dual-branch attention network for joint image deblurring
  and super-resolution.
\newblock In {\em CVPR Workshops}, 2021.

\bibitem{son2021ntire}
Sanghyun Son, Suyoung Lee, Seungjun Nah, Radu Timofte, Kyoung~Mu Lee, et~al.
\newblock {NTIRE 2021} challenge on video super-resolution.
\newblock In {\em CVPR Workshops}, 2021.

\bibitem{Su_2017_CVPR}
Shuochen Su, Mauricio Delbracio, Jue Wang, Guillermo Sapiro, Wolfgang Heidrich,
  and Oliver Wang.
\newblock Deep video deblurring for hand-held cameras.
\newblock In {\em CVPR}, 2017.

\bibitem{Suin_2020_CVPR}
Maitreya Suin, Kuldeep Purohit, and A.~N. Rajagopalan.
\newblock Spatially-attentive patch-hierarchical network for adaptive motion
  deblurring.
\newblock In {\em CVPR}, 2020.

\bibitem{svoboda2016compression}
Pavel Svoboda, Michal Hradis, David Barina, and Pavel Zemcik.
\newblock Compression artifacts removal using convolutional neural networks.
\newblock {\em arXiv preprint arXiv:1605.00366}, 2016.

\bibitem{Tao_2018_CVPR}
Xin Tao, Hongyun Gao, Xiaoyong Shen, Jue Wang, and Jiaya Jia.
\newblock Scale-recurrent network for deep image deblurring.
\newblock In {\em CVPR}, 2018.

\bibitem{Timofte_2017_CVPR_Workshops_ntire2017}
Radu Timofte, Eirikur Agustsson, Luc Van~Gool, Ming-Hsuan Yang, and Lei Zhang.
\newblock {NTIRE} 2017 challenge on single image super-resolution: Methods and
  results.
\newblock In {\em CVPR Workshops}, 2017.

\bibitem{Timofte_2016_CVPR}
Radu Timofte, Rasmus Rothe, and Luc Van~Gool.
\newblock Seven ways to improve example-based single image super resolution.
\newblock In {\em CVPR}, 2016.

\bibitem{Tong_2017_ICCV_srdensenet}
Tong Tong, Gen Li, Xiejie Liu, and Qinquan Gao.
\newblock Image super-resolution using dense skip connections.
\newblock In {\em ICCV}, 2017.

\bibitem{tsai2021banet}
Fu-Jen Tsai, Yan-Tsung Peng, Yen-Yu Lin, Chung-Chi Tsai, and Chia-Wen Lin.
\newblock {BAN}et: Blur-aware attention networks for dynamic scene deblurring.
\newblock {\em arXiv preprint arXiv:2101.07518}, 2021.

\bibitem{ulyanov2016instance}
Dmitry Ulyanov, Andrea Vedaldi, and Victor Lempitsky.
\newblock Instance normalization: The missing ingredient for fast stylization.
\newblock {\em arXiv preprint arXiv:1607.08022}, 2016.

\bibitem{vaswani2017attention}
Ashish Vaswani, Noam Shazeer, Niki Parmar, Jakob Uszkoreit, Llion Jones,
  Aidan~N Gomez, Lukasz Kaiser, and Illia Polosukhin.
\newblock Attention is all you need.
\newblock {\em arXiv preprint arXiv:1706.03762}, 2017.

\bibitem{Wang_2020_CVPR}
Li Wang, Dong Li, Yousong Zhu, Lu Tian, and Yi Shan.
\newblock Dual super-resolution learning for semantic segmentation.
\newblock In {\em CVPR}, 2020.

\bibitem{Wang_2019_CVPR_Workshops_edvr}
Xintao Wang, Kelvin~C.K. Chan, Ke Yu, Chao Dong, and Chen~Change Loy.
\newblock Edvr: Video restoration with enhanced deformable convolutional
  networks.
\newblock In {\em CVPR Workshops}, 2019.

\bibitem{Wang_2018_CVPR_non_local}
Xiaolong Wang, Ross Girshick, Abhinav Gupta, and Kaiming He.
\newblock Non-{L}ocal neural networks.
\newblock In {\em CVPR}, 2018.

\bibitem{Wang_2018_ECCV_Workshops_esrgan}
Xintao Wang, Ke Yu, Shixiang Wu, Jinjin Gu, Yihao Liu, Chao Dong, Yu Qiao, and
  Chen Change~Loy.
\newblock {ESRGAN}: Enhanced super-resolution generative adversarial networks.
\newblock In {\em ECCV Workshops}, 2018.

\bibitem{wang2004image_ssim}
Zhou Wang, Alan~C Bovik, Hamid~R Sheikh, Eero~P Simoncelli, et~al.
\newblock Image quality assessment: from error visibility to structural
  similarity.
\newblock {\em IEEE TIP}, 13(4):600--612, 2004.

\bibitem{Xu_2020_CVPR_Workshops}
Dejia Xu, Yihao Chu, and Qingyan Sun.
\newblock Moire pattern removal via attentive fractal network.
\newblock In {\em CVPR Workshops}, 2020.

\bibitem{xu2010two}
Li Xu and Jiaya Jia.
\newblock Two-phase kernel estimation for robust motion deblurring.
\newblock In {\em ECCV}, 2010.

\bibitem{Xu_2021_CVPR_Workshops_EDPN}
Ruikang Xu, Zeyu Xiao, Jie Huang, Yueyi Zhang, and Zhiwei Xiong.
\newblock {EDPN}: Enhanced deep pyramid network for blurry image restoration.
\newblock In {\em CVPR Workshops}, 2021.

\bibitem{Xu_2017_ICCV}
Xiangyu Xu, Deqing Sun, Jinshan Pan, Yujin Zhang, Hanspeter Pfister, and
  Ming-Hsuan Yang.
\newblock Learning to super-resolve blurry face and text images.
\newblock In {\em ICCV}, 2017.

\bibitem{yang2021ntire}
Ren Yang, Radu Timofte, et~al.
\newblock {NTIRE 2021} challenge on quality enhancement of compressed video:
  Methods and results.
\newblock In {\em CVPR Workshops}, 2021.

\bibitem{yang1995projection}
Yongyi Yang, Nikolas~P Galatsanos, and Aggelos~K Katsaggelos.
\newblock Projection-based spatially adaptive reconstruction of block-transform
  compressed images.
\newblock {\em IEEE TIP}, 4(7):896--908, 1995.

\bibitem{Yoo_2018_CVPR}
Jaeyoung Yoo, Sang-ho Lee, and Nojun Kwak.
\newblock Image restoration by estimating frequency distribution of local
  patches.
\newblock In {\em CVPR}, 2018.

\bibitem{yoo2014post}
Seok~Bong Yoo, Kyuha Choi, and Jong~Beom Ra.
\newblock Post-processing for blocking artifact reduction based on inter-block
  correlation.
\newblock {\em IEEE TMM}, 16(6):1536--1548, 2014.

\bibitem{yu2015multi}
Fisher Yu and Vladlen Koltun.
\newblock Multi-scale context aggregation by dilated convolutions.
\newblock {\em arXiv preprint arXiv:1511.07122}, 2015.

\bibitem{yu2018wide}
Jiahui Yu, Yuchen Fan, Jianchao Yang, Ning Xu, Zhaowen Wang, Xinchao Wang, and
  Thomas Huang.
\newblock Wide activation for efficient and accurate image super-resolution.
\newblock {\em arXiv preprint arXiv:1808.08718}, 2018.

\bibitem{Yu_2018_CVPR_rl_restore}
Ke Yu, Chao Dong, Liang Lin, and Chen~Change Loy.
\newblock Crafting a toolchain for image restoration by deep reinforcement
  learning.
\newblock In {\em CVPR}, 2018.

\bibitem{Yuan_2020_CVPR}
Yuan Yuan, Wei Su, and Dandan Ma.
\newblock Efficient dynamic scene deblurring using spatially variant
  deconvolution network with optical flow guided training.
\newblock In {\em CVPR}, 2020.

\bibitem{zamir2020learning}
Syed~Waqas Zamir, Aditya Arora, Salman Khan, Munawar Hayat, Fahad~Shahbaz Khan,
  Ming~Hsuan Yang, and Ling Shao.
\newblock Learning enriched features for real image restoration and
  enhancement.
\newblock In {\em ECCV}, 2020.

\bibitem{zamir2021multi}
Syed~Waqas Zamir, Aditya Arora, Salman Khan, Munawar Hayat, Fahad~Shahbaz Khan,
  Ming-Hsuan Yang, and Ling Shao.
\newblock Multi-stage progressive image restoration.
\newblock {\em arXiv preprint arXiv:2102.02808}, 2021.

\bibitem{Zhang_2019_CVPR_DMPHN}
Hongguang Zhang, Yuchao Dai, Hongdong Li, and Piotr Koniusz.
\newblock Deep stacked hierarchical multi-patch network for image deblurring.
\newblock In {\em CVPR}, 2019.

\bibitem{Zhang_2018_CVPR_svrn}
Jiawei Zhang, Jinshan Pan, Jimmy Ren, Yibing Song, Linchao Bao, Rynson~W.H.
  Lau, and Ming-Hsuan Yang.
\newblock Dynamic scene deblurring using spatially variant recurrent neural
  networks.
\newblock In {\em CVPR}, 2018.

\bibitem{Zhang_2019_CVPR_deep_plug}
Kai Zhang, Wangmeng Zuo, and Lei Zhang.
\newblock Deep plug-and-play super-resolution for arbitrary blur kernels.
\newblock In {\em CVPR}, 2019.

\bibitem{Zhang_2018_CVPR_unreasonable}
Richard Zhang, Phillip Isola, Alexei~A. Efros, Eli Shechtman, and Oliver Wang.
\newblock The unreasonable effectiveness of deep features as a perceptual
  metric.
\newblock In {\em CVPR}, 2018.

\bibitem{zhang2018gated}
Xinyi Zhang, Hang Dong, Zhe Hu, Wei-Sheng Lai, Fei Wang, and Ming-Hsuan Yang.
\newblock Gated fusion network for joint image deblurring and super-resolution.
\newblock In {\em BMVC}, 2018.

\bibitem{zhang2018deep}
Xinyi Zhang, Fei Wang, Hang Dong, and Yu Guo.
\newblock A deep encoder-decoder networks for joint deblurring and
  super-resolution.
\newblock In {\em ICASSP}, 2018.

\bibitem{Zhang_2018_ECCV_rcan}
Yulun Zhang, Kunpeng Li, Kai Li, Lichen Wang, Bineng Zhong, and Yun Fu.
\newblock Image super-resolution using very deep residual channel attention
  networks.
\newblock In {\em ECCV}, September 2018.

\bibitem{Zhang_2018_CVPR_rdn}
Yulun Zhang, Yapeng Tian, Yu Kong, Bineng Zhong, and Yun Fu.
\newblock Residual dense network for image super-resolution.
\newblock In {\em CVPR}, 2018.

\bibitem{zhao2020efficient}
Hengyuan Zhao, Xiangtao Kong, Jingwen He, Yu Qiao, and Chao Dong.
\newblock Efficient image super-resolution using pixel attention.
\newblock {\em arXiv preprint arXiv:2010.01073}, 2020.

\bibitem{zhong2020efficient_estrnn}
Zhihang Zhong, Ye Gao, Yinqiang Zheng, and Bo Zheng.
\newblock Efficient spatio-temporal recurrent neural network for video
  deblurring.
\newblock In {\em ECCV}, 2020.

\bibitem{Zhou_2019_ICCV_kmsr}
Ruofan Zhou and Sabine Susstrunk.
\newblock Kernel modeling super-resolution on real low-resolution images.
\newblock In {\em ICCV}, 2019.

\bibitem{Zhou_2019_CVPR_davanet}
Shangchen Zhou, Jiawei Zhang, Wangmeng Zuo, Haozhe Xie, Jinshan Pan, and
  Jimmy~S. Ren.
\newblock {DAVAN}et: Stereo deblurring with view aggregation.
\newblock In {\em CVPR}, 2019.

\bibitem{Zhu_2019_CVPR_deformable_V2}
Xizhou Zhu, Han Hu, Stephen Lin, and Jifeng Dai.
\newblock Deformable {C}onv{N}ets {V}2: More deformable, better results.
\newblock In {\em CVPR}, 2019.

\end{thebibliography}
}

\end{document}